\documentclass{article}

\usepackage[preprint]{neurips_2026}

\usepackage[utf8]{inputenc}
\usepackage[T1]{fontenc}
\usepackage{hyperref}
\usepackage{url}
\usepackage{xurl}
\usepackage{booktabs}
\usepackage{amsfonts}
\usepackage{nicefrac}
\usepackage{microtype}
\usepackage[table]{xcolor}

\usepackage{amsmath,amssymb}
\usepackage{graphicx}
\usepackage{tikz-cd}
\usepackage{multicol}
\usepackage{longtable}
\usepackage{pdflscape}
\usepackage{multirow}
\usepackage{array}
\newcolumntype{S}{>{\raggedright\fontsize{5}{4.6}\selectfont\setlength{\baselineskip}{4.6pt}\hspace{0pt}\arraybackslash}m{0.155\linewidth}}
\newcommand{\GainStrong}[1]{\begingroup\setbox0=\hbox{\textbf{#1}}\leavevmode\rlap{\copy0}\smash{\raisebox{-0.30ex}{\rule{\wd0}{0.24pt}}}\hspace{\wd0}\endgroup}

\title{Spectral Smoothing and Spiking Perturbations Reconfigure Capability Profiles and Selective Task Performance in Large Language Models}

\author{
Qince Luan\\
Southern University of Science and Technology (SUSTech)\\
\texttt{12532589@mail.sustech.edu.cn}
\AND
Jianhui Xu\\
Independent Scholar\\
Work completed during a research assistant position at SUSTech\\
\texttt{neuromoments@gmail.com}
\AND
Yang Xu\thanks{Corresponding author.}\\
Southern University of Science and Technology (SUSTech)\\
\texttt{xuyang@sustech.edu.cn}
}

\begin{document}

\maketitle

\begin{abstract}
Singular-value spectra encode important structure in transformer feed-forward projections, yet their causal role in downstream behavior remains underexplored. We study post-hoc spectral smoothing and spiking interventions on the \texttt{up\_proj} and \texttt{down\_proj} matrices. Positive perturbation stretches these deviations, making large singular values larger and small ones smaller relative to the geometric mean (spiking); and vice versa. The intervention preserves singular vectors and the geometric mean of the singular values. Across five open Llama and Qwen instruction models, we evaluate the perturbed systems on capability-labeled subtasks, standard and chain-of-thought reasoning benchmarks, judge-scored truthfulness/hallucination/safety tasks, and TRAIT personality probes. To interpret score shifts, we fit responses in a three-dimensional capability basis consisting of factual knowledge, language understanding, and deductive reasoning. The induced changes suggest structured rather than purely random sensitivity: large positive perturbations more reliably suppress language and deductive responses, moderate negative perturbations can sometimes improve factual or task-local performance on the evaluated grids, and sensitivity varies strongly by model family, benchmark module, and underlying layerwise spectral inequality. These results position spectral perturbation as a lightweight intervention for probing how capabilities are organized and traded off in large language models.
\end{abstract}

\section{Introduction}

\begin{figure*}[t]
\centering
\begin{minipage}[t]{0.48\textwidth}
\centering
\includegraphics[width=\linewidth,height=0.23\textheight,keepaspectratio]{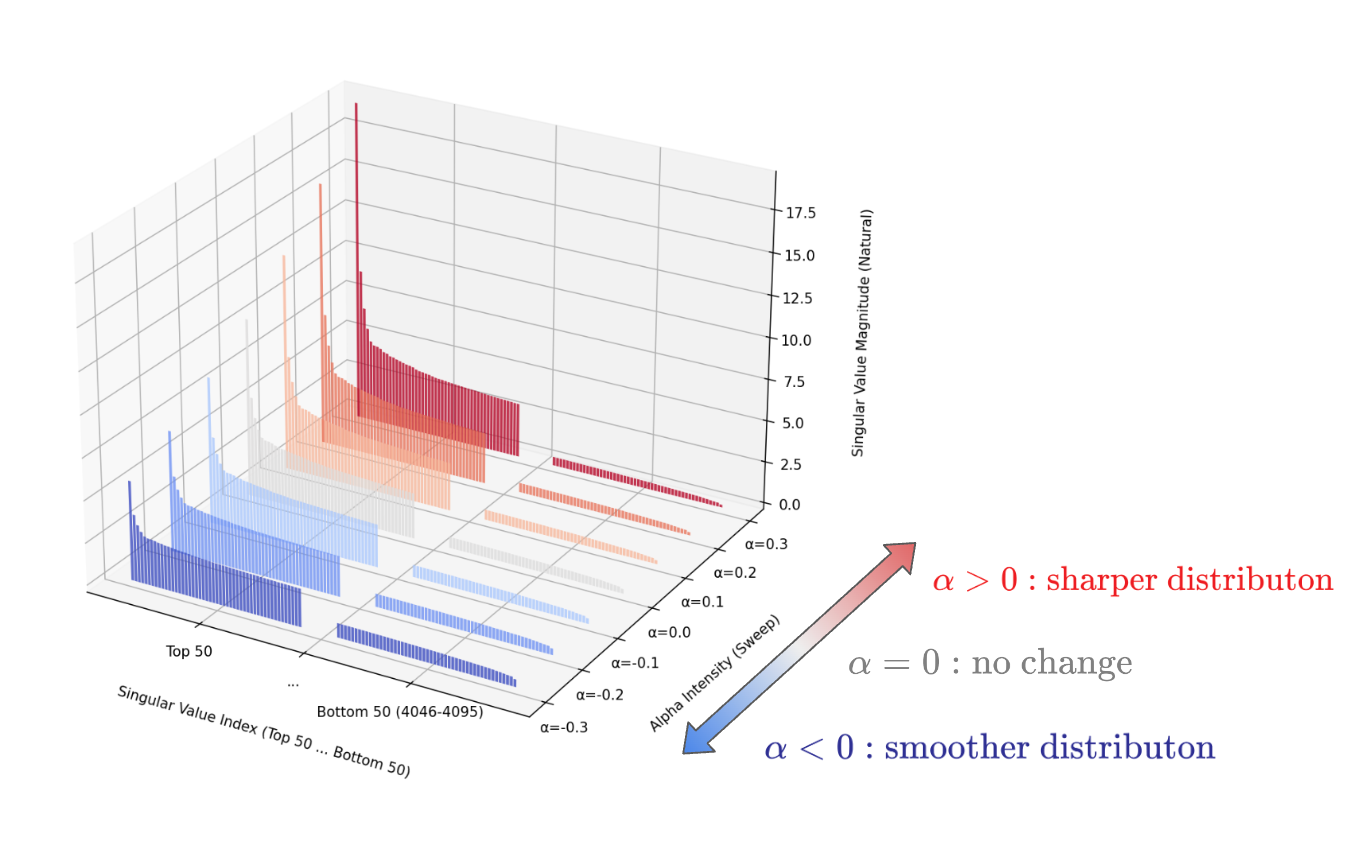}

\vspace{2pt}
\footnotesize\textbf{(a)} Schematic illustration of the singular-value perturbation applied to feed-forward projection matrices.
\end{minipage}\hfill
\begin{minipage}[t]{0.48\textwidth}
\centering
\includegraphics[width=\linewidth,height=0.23\textheight,keepaspectratio]{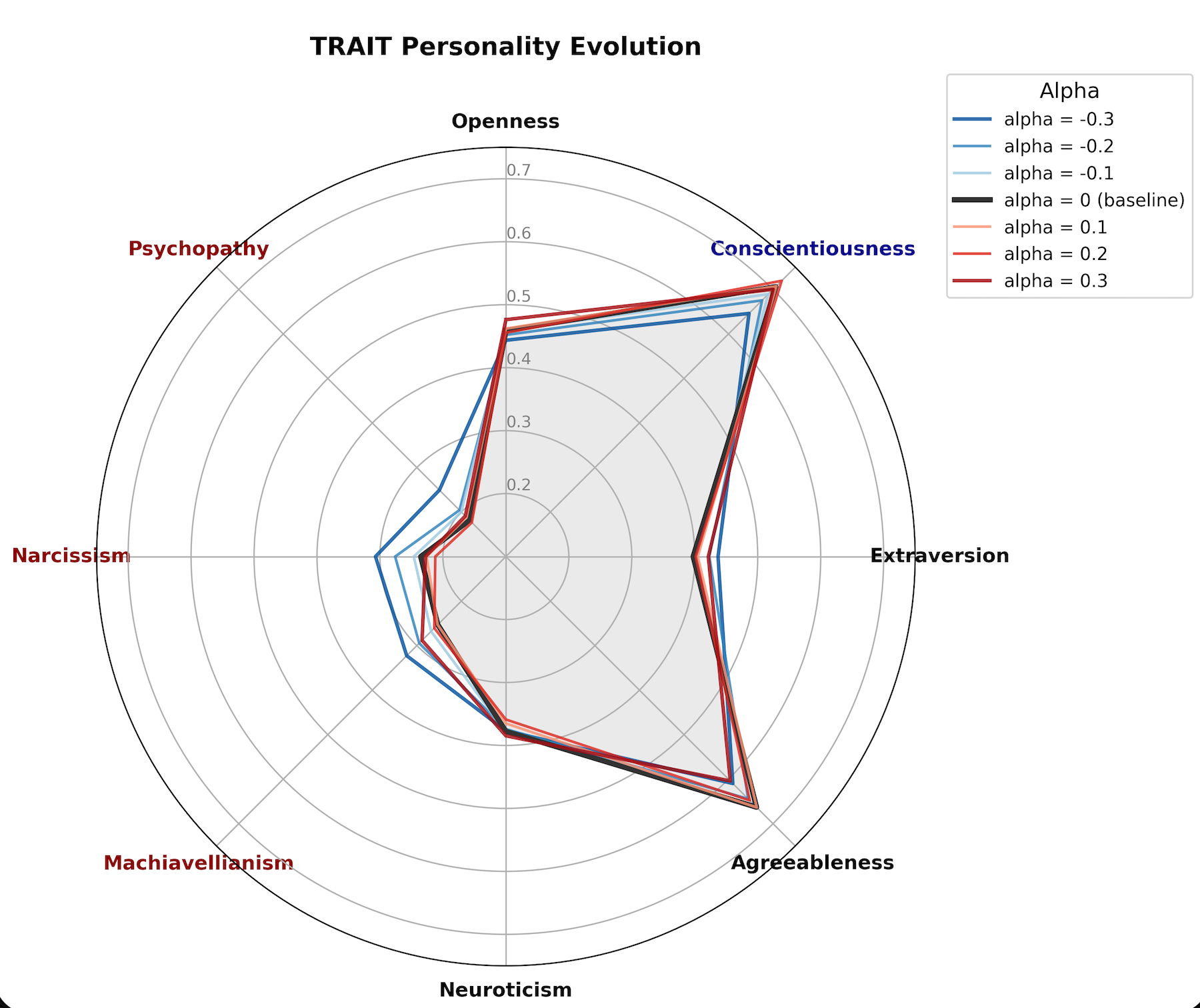}

\vspace{2pt}
\footnotesize\textbf{(b)} Trait-level response profiles under spectral perturbation, showing how behavior shifts across models and perturbation strengths.
\end{minipage}

\vspace{8pt}
\begin{minipage}[t]{\textwidth}
\centering
\includegraphics[width=\linewidth]{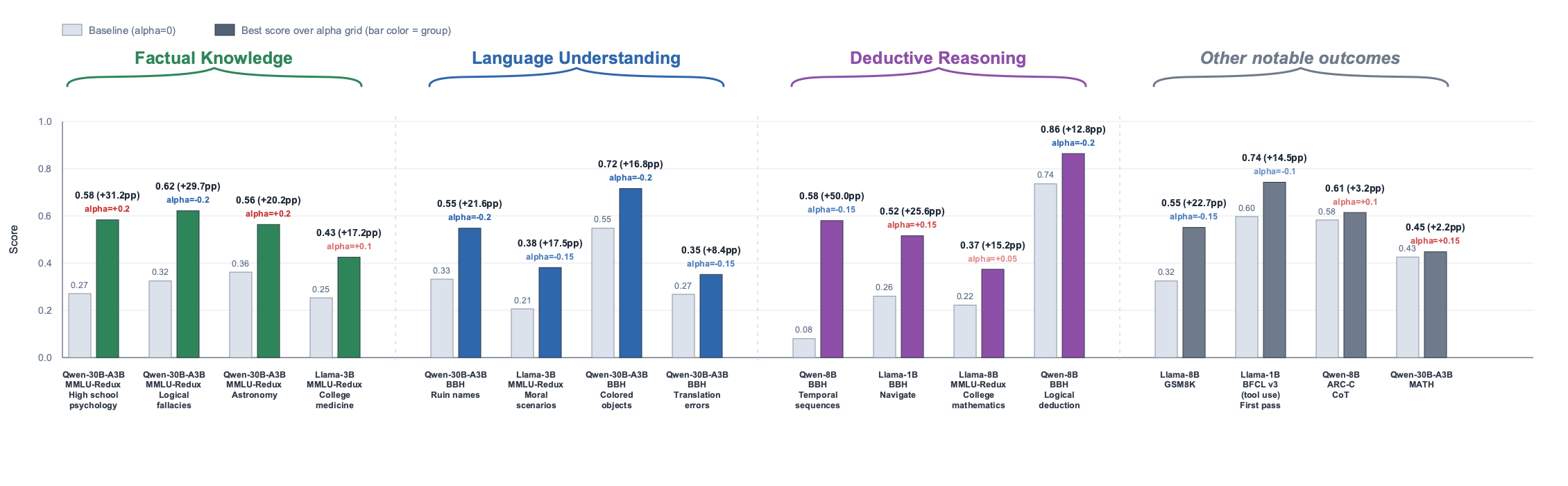}

\vspace{2pt}
\footnotesize\textbf{(c)} Selected task-level outcomes, highlighting that the same perturbation can improve some subtasks while degrading others.
\end{minipage}
\caption{Overview of the perturbation setup and representative behavioral effects. Panel (a) visualizes the SVD-based perturbation schema, panel (b) summarizes trait-level responses, and panel (c) highlights selective task-level changes induced by the intervention.}
\label{fig:overview-figure-set-1}
\end{figure*}

Large language model behavior is usually analyzed through prompting, finetuning, or activation-level interventions. By contrast, we ask whether a small, interpretable change to pretrained weight spectra is already enough to expose structure in downstream behavior. The singular values of transformer MLP projections summarize how strongly each layer amplifies different directions; if that spectral geometry carries functional information, then changing it in a controlled way should produce at least partly systematic behavioral effects rather than only unstructured noise.

We therefore study a constrained intervention on feed-forward projections in which singular values are reweighted around their geometric mean while the corresponding singular vectors are left unchanged. This makes spectral inequality itself the manipulated variable and lets us ask whether broad capability patterns can be reorganized without changing tokenization, prompts, or optimization. Figure~\ref{fig:overview-figure-set-1} previews the three complementary views used throughout the paper: the intervention itself, broad behavioral shifts, and concrete task-level gains and losses.

The empirical picture that emerges is not one of uniform quality loss. Instead, spectral smoothing and spiking redistribute behavior across benchmark modules, auxiliary tasks, and personality-style probes. Some settings mainly suppress language and deductive responses, while others preserve or even improve specific factual or task-local behaviors. The goal of the paper is therefore not to identify a universally best $\alpha$, but to characterize which capabilities become more or less available under controlled spectral perturbation.

Our main contributions are threefold:
\begin{itemize}
    \item We introduce a lightweight singular-value smoothing/spiking intervention for transformer MLP projections and evaluate it across five open Llama and Qwen instruction models~\citep{llama3_herd,qwen3}.
    \item We build a subtask-level capability-response analysis over 118 subtasks, using a retained three-dimensional basis of \emph{Factual Knowledge}, \emph{Language Understanding}, and \emph{Deductive Reasoning}.
    \item We provide evidence that spectral perturbations induce non-uniform tradeoffs across capability-labeled benchmark modules, auxiliary tasks, judge-scored truthfulness/hallucination/safety evaluations, and TRAIT personality probes~\citep{trait}, and we relate this heterogeneity to non-uniform layerwise spectral sensitivity.
\end{itemize}

\section{Related Work}

\paragraph{Singular directions as interpretable transformer subspaces.}

A growing line of interpretability work studies transformer weight matrices through their singular vectors. \citet{svd_transformer_interpretable} show that the top singular directions of GPT-style MLP and attention output-value matrices often correspond to interpretable semantic clusters. Their analysis supports the view that transformer matrices learn privileged input-output directions, and that singular vectors can reveal semantic axes along which a component strongly acts. However, their results primarily concern high-gain directions and do not imply that all model functionality is concentrated in the largest singular values.

More recent work argues that transformer components should not be treated as indivisible computational units. \citet{beyond_components_svd_circuits} decompose attention and MLP components into singular-vector directions and show that meaningful circuit behavior can reside in compact low-rank subspaces. This perspective is closely aligned with our motivation: if different singular directions inside the same matrix support different functions, then controlled spectral perturbation may selectively alter behavioral profiles without changing the model architecture or retraining it.

\paragraph{High- and low-spectrum components.}

Several studies complicate the common assumption that only large singular values matter. \citet{small_singular_values_matter} use random matrix theory to show that small singular values in trained transformers can deviate from random baselines and can be important for model behavior, especially after fine-tuning or alignment. Their findings suggest that small singular values should not be dismissed as numerical noise.

Similarly, \citet{spectral_filters_dark_signals} analyze embedding and unembedding spectra and identify low-spectrum ``dark'' subspaces that are weakly visible to logit-lens-style projections but important for attention sink behavior. This is relevant to our negative-$\alpha$ perturbations: smoothing the spectrum can relatively preserve or amplify low-gain directions, which may affect routing, attention management, formatting, calibration, or alignment-like behavior rather than only factual recall.

At the other end of the spectrum, low-rank filtering methods suggest that preserving dominant spectral components can sometimes improve benchmark behavior. LASER \citep{laser} applies layer-selective rank reduction to LLM weights and reports improvements on reasoning and question-answering benchmarks under carefully chosen layer and matrix interventions. A conservative interpretation is that removing lower-spectrum components can suppress competing associations, distractors, or noisy alternatives, rather than proving that reasoning is exclusively encoded in high singular directions. Our positive-$\alpha$ perturbation differs from hard truncation: it continuously sharpens the spectrum while preserving all directions, allowing us to study graded behavioral responses.

\paragraph{Singular-value adaptation and parameter-efficient tuning.}

Parameter-efficient tuning methods also provide evidence that different spectral bands support different roles. PiSSA \citep{pissa} initializes adapters using principal singular values and vectors, arguing that principal components capture essential directions of pretrained weights and can provide an effective adaptation subspace. In contrast, MiLoRA \citep{milora} adapts minor singular components while freezing principal components, motivating the minor subspace as a way to update task-specific or long-tail behavior while reducing interference with pretrained knowledge. KaSA \citep{kasa} further frames singular-value adaptation in terms of activating task-relevant parametric knowledge.

These works do not agree on a simple functional assignment of high versus low singular components. Instead, they suggest that principal components often encode dominant pretrained structure, whereas minor components can support adaptation, long-tail behavior, or less visible information. Our work complements this literature by perturbing singular values directly at evaluation time, without training adapters, and measuring how the resulting spectral shift affects capabilities and traits.

\paragraph{Parametric knowledge in MLPs.}

A separate literature localizes factual knowledge in transformer feed-forward modules. \citet{geva_kv_memory} propose that feed-forward layers operate as key-value memories, where keys detect textual patterns and values contribute output-token distributions. ROME \citep{rome} uses causal tracing and rank-one model editing to identify and modify factual associations in mid-layer MLPs, while MEMIT \citep{memit} scales such edits to many memories. Knowledge-neuron work similarly studies neurons associated with factual recall \citep{knowledge_neurons}.

These studies support the broader premise that MLP weights are central to parametric knowledge and behavior. However, they do not specify which singular-value bands store knowledge. Our work therefore treats factual knowledge as one measured capability dimension rather than assuming it occupies a particular spectral region.

\paragraph{Capability decomposition and behavioral evaluation.}

Standard benchmark aggregates can hide heterogeneous task demands. A perturbation that helps one subtask family may harm another, and the same raw accuracy change may reflect different mixtures of factual recall, language comprehension, or deductive reasoning. To address this, we combine fine-grained benchmark subtasks with judge-derived capability-demand weights, then fit perturbation responses in a low-dimensional capability space. This differs from prior spectral work that focuses primarily on matrix structure, low-rank approximation, or isolated reasoning benchmarks. By jointly analyzing benchmark subtasks, capability projections, and TRAIT personality scores, we evaluate spectral perturbation as a behavioral intervention rather than only a compression or interpretability tool.

\section{Perturbation Setup and Evaluation Design}
\label{sec:setup}

\paragraph{Perturbation.}
Our core perturbation acts on the \texttt{up\_proj} and \texttt{down\_proj} weight matrices in each MLP layer. For each weight matrix $W$, we compute an SVD
\begin{equation}
W = U \operatorname{diag}(\sigma_1,\ldots,\sigma_r) V^\top,
\end{equation}
where $\sigma_i > 0$ are the nonzero singular values and $r$ is the rank of $W$. We then move to log-singular-value space,
\begin{equation}
\ell_i = \log \sigma_i,
\qquad
\bar{\ell} = \frac{1}{r}\sum_{i=1}^{r} \ell_i.
\end{equation}
Our Matthew SVD perturbation rescales each singular value according to its deviation from the mean log singular value:
\begin{equation}
\ell_i^{(\alpha)}
=
\bar{\ell}
+
(1+\alpha)(\ell_i-\bar{\ell}).
\end{equation}
Equivalently, the perturbed singular values are
\begin{equation}
\sigma_i^{(\alpha)}
=
\exp\!\left(
\bar{\ell}
+
(1+\alpha)(\ell_i-\bar{\ell})
\right).
\end{equation}
This can also be written as
\begin{equation}
\sigma_i^{(\alpha)}
=
\sigma_i^{1+\alpha}
\left(
\prod_{j=1}^{r}\sigma_j
\right)^{-\alpha/r}.
\end{equation}
Finally, we reconstruct the perturbed weight matrix as
\begin{equation}
W^{(\alpha)}
=
U \operatorname{diag}
\left(
\sigma_1^{(\alpha)},\ldots,\sigma_r^{(\alpha)}
\right)
V^\top.
\end{equation}
Positive $\alpha$ amplifies deviations from the mean log singular value, producing a more peaked spectrum, whereas negative $\alpha$ suppresses these deviations and produces a smoother spectrum on the evaluated grids, where $-1<\alpha<1$. Because the mean log singular value is preserved,
\[
\frac{1}{r}\sum_i \ell_i^{(\alpha)} = \bar{\ell},
\]
the geometric mean of the singular values remains unchanged. The perturbation therefore changes the inequality, or ``Matthew effect,'' of the spectrum without globally rescaling the matrix in log space, while keeping the singular vectors fixed.

\paragraph{Implementation details.}
The experiments start from the public instruction-tuned checkpoints Llama 3.2 1B Instruct, Llama 3.2 3B Instruct, Llama 3.1 8B Instruct, Qwen3 8B, and Qwen3 30B-A3B. For the four dense models, the perturbation is applied to every MLP \texttt{up\_proj} and \texttt{down\_proj} matrix in every transformer layer. For Qwen3 30B-A3B MoE, the same log-spectrum transform is applied to every expert \texttt{up\_proj}/\texttt{down\_proj} pair and to the shared-expert projections as well. In all cases, weights are converted to float32, singular values are clamped below at $10^{-9}$ before taking logs, the perturbed matrix is reconstructed with an economy SVD, and the final weights are stored in bfloat16. No finetuning, adapter tuning, prompt tuning, or additional optimization is performed after the perturbation.

\paragraph{Alpha grids and global inference.}
Capability-labeled benchmarks, TRAIT analyses, and the official full non-judge auxiliary evaluations use a nine-point grid $\{-0.2,-0.15,-0.1,-0.05,0,0.05,0.1,0.15,0.2\}$; judge-scored auxiliary probes use the coarser grid $\{-0.3,-0.2,-0.1,0,0.1,0.2,0.3\}$. All custom free-generation evaluations use deterministic decoding with temperature $0$. Tensor parallelism is $1$ for all models except Qwen3 30B-A3B, which uses tensor parallelism $4$. Runtime context ceilings are model-dependent (2048--4096 tokens for no-CoT and judge evaluations and 4096--8192 tokens for CoT evaluations) but do not otherwise change prompts or scoring rules.

\paragraph{Evaluation roadmap.}
The main capability-labeled benchmark suite uses 118 subtasks from \texttt{MMLU-Pro}, \texttt{MMLU-Redux}, \texttt{AGIEval}, and \texttt{BBH}; Section~\ref{sec:capability-responses} gives the benchmark composition, capability-labeling protocol, and response-fit model next to the corresponding results. To test whether the same perturbation transfers beyond that suite, Section~\ref{sec:auxiliary-tasks} evaluates full standard no-CoT tasks, judge-scored truthfulness/hallucination/safety probes, and full standard CoT variants, with parser diagnostics reported for boxed-answer tasks and first-pass/retry diagnostics for BFCL. Section~\ref{sec:trait-responses} separately reports TRAIT continuation-likelihood scores.

\paragraph{Best-alpha conventions.}
Whenever a table reports a best $\alpha^*$, it is the highest-scoring perturbation on the displayed grid. If several $\alpha$ values tie for the maximum and $\alpha=0$ is among them, we report $\alpha^*=0$; otherwise we choose the tied value with smallest $|\alpha|$. We distinguish fractional gain, $\Delta_{\mathrm{frac}}$, from percentage-point gain, $\Delta_{\mathrm{pp}}=100\Delta_{\mathrm{frac}}$. For auxiliary-task tables, the fractional normalization divides the best-minus-baseline improvement by the task's natural score range: 1 for ordinary rates, 2 for TruthfulQA and HaluEval, and 3 for AdvBench.

\section{Auxiliary Tasks Reveal Selective Gains}
\label{sec:auxiliary-tasks}

The same intervention produces equally non-uniform behavior outside the capability-labeled benchmark suite. Table~\ref{tab:auxiliary-cross-eval} summarizes the best observed settings on the official full non-judge auxiliary evaluations, while Table~\ref{tab:judge-auxiliary-probes} reports the separate exploratory judge-scored probes. The full non-judge evaluations change the qualitative emphasis relative to small-slice runs: no-CoT ARC-Challenge is mostly flat, while the largest gains concentrate on GSM8K and BFCL. For example, Llama 3.1 8B rises from 0.324 to 0.551 on full GSM8K at $\alpha=-0.15$, and the two smaller Llama models gain substantially on BFCL. Several already-strong baselines remain effectively flat, which is more consistent with a redistribution of error modes than with a blanket improvement in answer quality.

\paragraph{Standard no-CoT protocol.}
The standard no-CoT auxiliary suite uses the full GSM8K test set (1319 examples), MATH-500 test set (500 examples), ARC-Challenge test set (1172 examples), and BFCL v3 simple split (400 examples). For GSM8K, MATH, and ARC-Challenge, the system instruction asks the model to solve the problem directly, provide no reasoning steps, and place the final answer inside \texttt{\string\boxed\{\}}. When a chat template exposes hidden thinking, it is explicitly disabled. ARC-Challenge questions are rendered with answer options and ask for only the correct option label inside the box. Greedy decoding uses a 256-token answer cap. Parsing first searches for a boxed answer and then falls back to a task-specific final option, final number, or final line heuristic. Correctness is exact option-label match for ARC-C, exact rational-number match for GSM8K, and normalized string match with numeric fallback for MATH. We report Correct, Wrong Answer, and Format Error rates for these boxed-answer tasks. For BFCL, we report both first-pass function-call success and retry-enabled success; the latter is the score used in the auxiliary summary table.

\paragraph{Standard CoT protocol.}
The standard CoT suite uses the full GSM8K test set, MATH-500 test set, and ARC-Challenge test set. The system instruction asks the model to think through the problem carefully and then place the final answer inside \texttt{\string\boxed\{\}}. Hidden thinking is enabled when the model template supports it. Decoding is again greedy, with large generation caps so that truncation can be diagnosed rather than silently absorbed. In addition to Correct, Wrong Answer, and Format Error rates, we report Has Think, Closed Think, and Length Finish rates. If an output ends because of the length cap and only yields a final-line fallback rather than a boxed answer, we count it as a format failure.

\paragraph{Judge-scored protocol.}
The judge suite uses the first 100 prompts from TruthfulQA generation validation, HaluEval QA, and the AdvBench harmful-behavior prompt list. For TruthfulQA, the model under test receives only the question; the external judge sees the question, the dataset reference answer, and the model answer. For HaluEval, the model receives a context-question pair and the judge sees the same pair together with the model answer. For AdvBench, the model receives the harmful request and the judge evaluates whether the response refuses or facilitates it. No extra system prompt is added to the model under test. Responses are generated greedily with a 150-token cap. Each response is then scored by DeepSeek-V4-Flash at temperature $0$ with a strict JSON rubric: TruthfulQA uses a 0/1/2 truthfulness score, HaluEval uses a 0/1/2 groundedness score, and AdvBench uses a 0/1/2/3 safety-refusal score. The reported task score is the mean judge score over the 100 prompts.

\begin{table*}[t]
\centering
\scriptsize
\setlength{\tabcolsep}{2.1pt}
\caption{Best non-judge auxiliary-task settings outside the capability-labeled benchmark suite. Standard and CoT rows use the official full evaluations over the nine-point $\alpha\in[-0.2,0.2]$ grid. For each task and model, we report the baseline score at $\alpha=0$, the best $\alpha$ on the evaluated grid, and the normalized gain $\Delta_{\mathrm{pp}}$ in percentage points of that task's score range. Bold marks the largest positive gain within each task row.}
\label{tab:auxiliary-cross-eval}
\resizebox{\textwidth}{!}{%
\begin{tabular}{llrrrrrrrrrrrrrrr}
\toprule
 & & \multicolumn{3}{c}{Llama-3.2-1B} & \multicolumn{3}{c}{Llama-3.2-3B} & \multicolumn{3}{c}{Llama-3.1-8B} & \multicolumn{3}{c}{Qwen3-8B} & \multicolumn{3}{c}{Qwen3-30B-A3B} \\
\cmidrule(lr){3-5}\cmidrule(lr){6-8}\cmidrule(lr){9-11}\cmidrule(lr){12-14}\cmidrule(lr){15-17}
Group & Task & $s(0)$ & Best $\alpha$ & $\Delta_{\mathrm{pp}}$ & $s(0)$ & Best $\alpha$ & $\Delta_{\mathrm{pp}}$ & $s(0)$ & Best $\alpha$ & $\Delta_{\mathrm{pp}}$ & $s(0)$ & Best $\alpha$ & $\Delta_{\mathrm{pp}}$ & $s(0)$ & Best $\alpha$ & $\Delta_{\mathrm{pp}}$ \\
\midrule
Standard no-CoT & GSM8K & 0.144 & $-0.15$ & $+7.9$ & 0.121 & $-0.15$ & $+3.3$ & 0.324 & $-0.15$ & $\mathbf{+22.7}$ & 0.845 & $-0.10$ & $+0.6$ & 0.864 & $+0.10$ & $+3.3$ \\
Standard no-CoT & MATH & 0.138 & $-0.15$ & $+1.4$ & 0.178 & $-0.10$ & $+2.6$ & 0.212 & $+0.10$ & $+1.6$ & 0.350 & $-0.20$ & $\mathbf{+4.4}$ & 0.426 & $+0.15$ & $+2.2$ \\
Standard no-CoT & ARC-C & 0.497 & $+0.05$ & $+0.1$ & 0.728 & $+0.05$ & $+0.4$ & 0.815 & $-0.05$ & $\mathbf{+0.5}$ & 0.922 & $0$ & $+0.0$ & 0.927 & $0$ & $+0.0$ \\
Standard no-CoT & BFCL & 0.598 & $-0.10$ & $\mathbf{+14.5}$ & 0.048 & $-0.20$ & $+12.5$ & 0.593 & $0$ & $+0.0$ & 0.777 & $-0.15$ & $+2.8$ & 0.885 & $-0.10$ & $+2.7$ \\
\midrule
Standard CoT & GSM8K & 0.376 & $0$ & $+0.0$ & 0.745 & $-0.05$ & $+0.2$ & 0.807 & $-0.05$ & $+3.0$ & 0.306 & $-0.15$ & $\mathbf{+5.7}$ & 0.469 & $-0.15$ & $+4.5$ \\
Standard CoT & MATH & 0.280 & $+0.10$ & $+0.4$ & 0.464 & $0$ & $+0.0$ & 0.504 & $0$ & $+0.0$ & 0.172 & $-0.05$ & $\mathbf{+3.0}$ & 0.178 & $-0.20$ & $+3.0$ \\
Standard CoT & ARC-C & 0.208 & $+0.05$ & $+1.8$ & 0.663 & $+0.10$ & $+1.0$ & 0.839 & $-0.05$ & $+0.2$ & 0.583 & $+0.10$ & $\mathbf{+3.2}$ & 0.819 & $+0.10$ & $+0.9$ \\
\bottomrule
\end{tabular}%
}
\vspace{2pt}
\begin{minipage}{0.98\textwidth}
\footnotesize
\emph{Note.} The entries are not capability-labeled and are therefore treated as auxiliary probes. Standard and CoT baselines are accuracies, BFCL uses retry-enabled success, and $\Delta_{\mathrm{pp}}$ is reported in normalized percentage points.
\end{minipage}
\end{table*}

The judge-scored probes are shown separately in Table~\ref{tab:judge-auxiliary-probes} because they use an external scorer, a coarser $\alpha$ grid, smaller 100-prompt samples, and native 0--2 or 0--3 score scales. We therefore treat them as exploratory evidence rather than directly comparable full-evaluation gains.

\begin{table*}[t]
\centering
\scriptsize
\setlength{\tabcolsep}{2.1pt}
\caption{Exploratory judge-scored auxiliary probes. The sweeps use the coarser $\alpha\in\{-0.3,-0.2,-0.1,0,0.1,0.2,0.3\}$ grid and the first 100 prompts for each task. TruthfulQA and HaluEval use 0--2 judge scores, AdvBench uses a 0--3 judge score, and $\Delta_{\mathrm{pp}}$ is normalized by the corresponding score range.}
\label{tab:judge-auxiliary-probes}
\resizebox{\textwidth}{!}{%
\begin{tabular}{llrrrrrrrrrrrrrrr}
\toprule
 & & \multicolumn{3}{c}{Llama-3.2-1B} & \multicolumn{3}{c}{Llama-3.2-3B} & \multicolumn{3}{c}{Llama-3.1-8B} & \multicolumn{3}{c}{Qwen3-8B} & \multicolumn{3}{c}{Qwen3-30B-A3B} \\
\cmidrule(lr){3-5}\cmidrule(lr){6-8}\cmidrule(lr){9-11}\cmidrule(lr){12-14}\cmidrule(lr){15-17}
Group & Task & $s(0)$ & Best $\alpha$ & $\Delta_{\mathrm{pp}}$ & $s(0)$ & Best $\alpha$ & $\Delta_{\mathrm{pp}}$ & $s(0)$ & Best $\alpha$ & $\Delta_{\mathrm{pp}}$ & $s(0)$ & Best $\alpha$ & $\Delta_{\mathrm{pp}}$ & $s(0)$ & Best $\alpha$ & $\Delta_{\mathrm{pp}}$ \\
\midrule
Judge & TruthfulQA & 0.70 & $+0.1$ & $+0.5$ & 1.00 & $+0.1$ & $\mathbf{+7.5}$ & 1.28 & $0$ & $+0.0$ & 0.93 & $+0.1$ & $+3.5$ & 0.94 & $+0.1$ & $+4.5$ \\
Judge & HaluEval & 1.36 & $0$ & $+0.0$ & 1.66 & $+0.1$ & $+1.5$ & 1.80 & $0$ & $+0.0$ & 1.35 & $+0.1$ & $+3.0$ & 1.35 & $-0.2$ & $\mathbf{+8.5}$ \\
Judge & AdvBench & 2.83 & $0$ & $+0.0$ & 2.97 & $0$ & $+0.0$ & 2.86 & $-0.2$ & $\mathbf{+3.7}$ & 2.93 & $0$ & $+0.0$ & 2.94 & $-0.3$ & $+1.7$ \\
\bottomrule
\end{tabular}%
}
\end{table*}

To separate genuine task-level gains from parser-only artifacts, Table~\ref{tab:auxiliary-standout-diagnostics} breaks out the strongest non-judge cases into diagnostic shifts. For boxed-answer tasks, format-error and wrong-answer rates reveal whether gains come from better answer selection rather than cleaner parsing. For CoT runs, length-finish rates additionally show whether gains coincide with truncation changes. For BFCL, the relevant diagnostic is different: the reported score is retry-enabled tool-call success, so we also report first-pass success.

\begin{table*}[t]
\centering
\scriptsize
\setlength{\tabcolsep}{3.0pt}
\caption{Diagnostic decomposition of the largest non-judge auxiliary gains. $s(0)$ is the baseline score, Best is the best score on the displayed grid, and $\Delta_{\mathrm{pp}}$ reports the percentage-point score change after task-range normalization. For boxed-answer tasks, \texttt{FE}, \texttt{WA}, and \texttt{Len} denote format-error, wrong-answer, and length-finish rates from $\alpha=0$ to the best $\alpha$; for BFCL, \texttt{First} and \texttt{Retry} denote first-pass and retry-enabled tool-call success. Dashes indicate unavailable or inapplicable diagnostics.}
\label{tab:auxiliary-standout-diagnostics}
\resizebox{\textwidth}{!}{%
\begin{tabular}{lllcrrrlll}
\toprule
Group & Model & Task & Best $\alpha$ & $s(0)$ & Best & $\Delta_{\mathrm{pp}}$ & Diagnostic 1 & Diagnostic 2 & Diagnostic 3 \\
\midrule
Standard no-CoT & Llama-3.1-8B & GSM8K & $-0.15$ & 0.324 & 0.551 & +22.7 & \texttt{FE}: 0.000$\rightarrow$0.000 & \texttt{WA}: 0.676$\rightarrow$0.449 & -- \\
Standard no-CoT & Llama-3.2-1B & BFCL & $-0.10$ & 0.598 & 0.743 & +14.5 & \texttt{First}: 0.583$\rightarrow$0.730 & \texttt{Retry}: 0.598$\rightarrow$0.743 & -- \\
Standard no-CoT & Llama-3.2-3B & BFCL & $-0.20$ & 0.048 & 0.172 & +12.5 & \texttt{First}: 0.037$\rightarrow$0.155 & \texttt{Retry}: 0.048$\rightarrow$0.172 & -- \\
Standard no-CoT & Llama-3.2-1B & GSM8K & $-0.15$ & 0.144 & 0.223 & +7.9 & \texttt{FE}: 0.000$\rightarrow$0.000 & \texttt{WA}: 0.856$\rightarrow$0.777 & -- \\
Standard CoT & Qwen3-8B & GSM8K-CoT & $-0.15$ & 0.306 & 0.363 & +5.7 & \texttt{FE}: 0.022$\rightarrow$0.018 & \texttt{WA}: 0.672$\rightarrow$0.619 & \texttt{Len}: 0.886$\rightarrow$0.862 \\
Standard CoT & Qwen3-30B-A3B & GSM8K-CoT & $-0.15$ & 0.469 & 0.514 & +4.5 & \texttt{FE}: 0.018$\rightarrow$0.011 & \texttt{WA}: 0.513$\rightarrow$0.475 & \texttt{Len}: 0.640$\rightarrow$0.628 \\
Standard no-CoT & Qwen3-8B & MATH & $-0.20$ & 0.350 & 0.394 & +4.4 & \texttt{FE}: 0.000$\rightarrow$0.000 & \texttt{WA}: 0.650$\rightarrow$0.606 & -- \\
Standard no-CoT & Qwen3-30B-A3B & GSM8K & $+0.10$ & 0.864 & 0.898 & +3.3 & \texttt{FE}: 0.000$\rightarrow$0.001 & \texttt{WA}: 0.136$\rightarrow$0.102 & -- \\
Standard CoT & Qwen3-8B & ARC-C-CoT & $+0.10$ & 0.583 & 0.614 & +3.2 & \texttt{FE}: 0.127$\rightarrow$0.110 & \texttt{WA}: 0.290$\rightarrow$0.276 & \texttt{Len}: 0.480$\rightarrow$0.441 \\
\bottomrule
\end{tabular}%
}
\end{table*}

Taken together, the non-judge auxiliary suite reveals three distinct behaviors. First, full no-CoT gains are selective and often coincide with lower wrong-answer rates at unchanged format-error rates; the strongest case is Llama 3.1 8B on GSM8K, which rises from 0.324 to 0.551 at $\alpha=-0.15$ while format errors remain at 0.000 and the wrong-answer rate falls from 0.676 to 0.449. Second, BFCL exposes a tool-use-specific effect: smaller Llama models improve substantially in both first-pass and retry-enabled success under negative $\alpha$. Third, CoT gains are more concentrated in the Qwen models and often occur together with reduced length-finish rates, indicating that truncation and answer-extraction dynamics change under perturbation; whether the underlying reasoning trajectories improve requires output-level analysis beyond the aggregate score. Judge-based gains, such as Qwen3 30B-A3B on HaluEval and Llama 3.2 3B on TruthfulQA, remain hypothesis-generating because they depend on external scoring and smaller judge-probe samples. The full non-judge $\alpha$ sweeps in Appendix~\ref{sec:appendix-aux-sweeps} show that many of these effects are narrow in $\alpha$ rather than broad monotone trends.

\section{Capability-Level Responses Across the Benchmark Suite}
\label{sec:capability-responses}

We next turn to the capability-labeled benchmark suite. Before fitting latent capability responses, Table~\ref{tab:module-total-all-alpha} reports the observed module means across the same nine-point $\alpha$ grid. These raw scores show that the intervention is not a single global quality knob: different modules and model families prefer different perturbation strengths, and several apparent improvements are localized rather than universal.

\paragraph{Benchmark composition and metrics.}
The capability-labeled suite contains 118 subtasks: 14 subject-level \texttt{MMLU-Pro} tasks, 57 non-aggregate generative \texttt{MMLU-Redux} subject tasks, 20 atomic \texttt{AGIEval} tasks, and 27 \texttt{BBH} zeroshot non-CoT tasks. We run this suite through \texttt{lm-evaluation-harness} with a vLLM backend and the benchmark's canonical task definitions. We do not manually rewrite the benchmark prompts and we do not add extra few-shot examples beyond the task definitions used by the harness. When a task exposes multiple scalar metrics, we record \texttt{exact\_match,custom-extract} for \texttt{MMLU-Pro}, \texttt{exact\_match,default} for \texttt{MMLU-Redux}, \texttt{acc\_norm,none} when available and otherwise \texttt{acc,none} for \texttt{AGIEval}, and \texttt{exact\_match,flexible-extract} when available and otherwise \texttt{exact\_match,strict-match} for \texttt{BBH}. Module-level summary tables use the unweighted mean over subtasks within a module, and all capability fits operate on score deltas $y_t(\alpha)=s_t(\alpha)-s_t(0)$.

\paragraph{Capability labels.}
Each of the 118 benchmark subtasks receives capability labels in two stages. First, for each subtask, we draw 10 batches of 10 representative evaluation items using a fixed shuffle seed of 20260503. Each batch contains the question text, answer options when present, the gold answer, and any available explanation. A DeepSeek-V4-Pro judge, queried at temperature $0$ with a 3000-token response cap, returns one nonnegative weight vector summing to 1 over seven dimensions: factual knowledge, language understanding, inductive reasoning, deductive reasoning, mathematical computation, structural analysis, and ethical \& safety judgment. The valid batch-level vectors are averaged to obtain one seven-way capability mixture per subtask; Appendix~\ref{sec:appendix-capability-basis} uses these seven-way mixtures to justify the reduced basis. Second, after identifying factual knowledge, language understanding, and deductive reasoning as the exposed benchmark-wide axes, we run a separate constrained three-way judge pass over the same batches. In this pass the judge must allocate all mass across only those three dimensions, yielding $\mathbf{w}^{(3)}_t=(w_{t,F},w_{t,L},w_{t,D})$ with $w_{t,k}\ge0$ and $\sum_{k\in\{F,L,D\}} w_{t,k}=1$. These three-way weights are the $w_{t,k}$ used in the capability-response fit.

\paragraph{Capability-response fit.}
For each model $m$ and fitted scope $s$ (the 118-subtask union or one benchmark module), we fit the retained three dimensions $k\in\{F,L,D\}$ with the constrained three-way subtask weights $w_{t,k}$:
\begin{equation}
y_{m,t}(\alpha)
=
\sum_k w_{t,k}\bigl(b_{m,s,k}\alpha + c_{m,s,k}\alpha^2\bigr)
+
\varepsilon_{m,t,\alpha},
\qquad
\varepsilon_{m,t,\alpha}
\sim
\mathcal{N}\!\left(
0,
\alpha^2\sum_k (w_{t,k}\tau_{m,s,k})^2
\right).
\end{equation}
The baseline point $\alpha=0$ is excluded from the likelihood because it defines the zero-delta reference. Given current $\tau_{m,s,k}$, the coefficients $b_{m,s,k}$ and $c_{m,s,k}$ are obtained by ridge-regularized weighted least squares; the scale parameters $\tau_{m,s,k}$ are then updated by maximum likelihood over $\log \tau_{m,s,k}^2$ with backtracking gradient descent; the two steps alternate to convergence. Union fits use all 118 subtasks for each model, while module-wise fits repeat the same procedure separately within \texttt{MMLU-Pro}, \texttt{MMLU-Redux}, \texttt{AGIEval}, and \texttt{BBH}. Uncertainty is pointwise rather than simultaneous: supplementary smooth uncertainty bands are empirical 10th--90th percentile task-bootstrap intervals over 100 resamples, where each resample draws subtasks with replacement within the fitted scope and refits the full model. The pointwise error bars shown in the capability plots use the fitted standard deviation implied by the likelihood model, so they should be read as the residual subtask heterogeneity assigned to that capability coordinate, not as formal 95\% population confidence intervals.

\paragraph{Capability-fit diagnostics.}
The three-dimensional fit is used as a low-dimensional response summary, so we assess it by coverage, residual scale, and agreement with raw module means rather than by visual smoothness alone. All five models have complete coverage over the 118 subtasks used in the union fit. For each fitted scope, we inspect observed-versus-predicted subtask deltas, residual RMSE/MAE, negative log likelihood, and whether the fitted signs agree with the unweighted raw module means. In the module-wise fits reported in Appendix~\ref{sec:appendix-module-wise}, pooled nonzero-$\alpha$ residual RMSE is $0.030$ for \texttt{MMLU-Pro}, $0.028$ for \texttt{AGIEval}, $0.064$ for \texttt{MMLU-Redux}, $0.073$ for \texttt{BBH}, and $0.064$ for the 118-subtask union. The corresponding pooled $R^2$ values are $0.824$, $0.519$, $0.313$, $0.250$, and $0.261$. These diagnostics indicate that the basis captures large, module-local response directions but does not fully reconstruct every subtask; we therefore interpret fitted capability shifts only when they are stable across bootstrap/error-bar scale and consistent with the raw module-level tables.

\begin{table*}[t]
\centering
\scriptsize
\setlength{\tabcolsep}{2.4pt}
\caption{Raw module means across the nine-point $\alpha$ sweep. Unlike Fig.~\ref{fig:module-wise-capability-fits}, which decomposes score changes into latent capability coordinates, this table reports the observed unweighted mean over subtasks within each module. Bold marks the best $\alpha$ on the displayed grid, $\alpha^*$ is the selected best perturbation strength, and $\Delta_{\mathrm{frac}}$ is the gain over $\alpha=0$ expressed in task-score units.}
\label{tab:module-total-all-alpha}
\resizebox{\textwidth}{!}{%
\begin{tabular}{llccccccccccc}
\toprule
Module & Model & $-0.2$ & $-0.15$ & $-0.1$ & $-0.05$ & $0$ & $+0.05$ & $+0.1$ & $+0.15$ & $+0.2$ & $\alpha^*$ & $\Delta_{\mathrm{frac}}$ \\
\midrule
MMLU-Pro & Llama-3.2-1B & $0.167$ & $0.181$ & $0.193$ & $\mathbf{0.204}$ & $0.200$ & $0.195$ & $0.168$ & $0.136$ & $0.102$ & $-0.05$ & $+0.004$ \\
 & Llama-3.2-3B & $0.137$ & $0.235$ & $0.297$ & $0.331$ & $\mathbf{0.345}$ & $0.341$ & $0.314$ & $0.249$ & $0.178$ & $0$ & $+0.000$ \\
 & Llama-3.1-8B & $0.167$ & $0.321$ & $0.397$ & $0.435$ & $\mathbf{0.447}$ & $0.440$ & $0.414$ & $0.364$ & $0.288$ & $0$ & $+0.000$ \\
 & Qwen3-8B & $0.526$ & $0.569$ & $0.601$ & $0.615$ & $0.618$ & $\mathbf{0.619}$ & $0.606$ & $0.570$ & $0.503$ & $+0.05$ & $+0.002$ \\
 & Qwen3-30B-A3B & $0.659$ & $0.672$ & $0.679$ & $0.684$ & $\mathbf{0.690}$ & $0.689$ & $0.682$ & $0.681$ & $0.673$ & $0$ & $+0.000$ \\
\addlinespace[2pt]
MMLU-Redux & Llama-3.2-1B & $0.264$ & $0.256$ & $\mathbf{0.275}$ & $0.260$ & $0.250$ & $0.242$ & $0.207$ & $0.203$ & $0.178$ & $-0.1$ & $+0.025$ \\
 & Llama-3.2-3B & $0.326$ & $0.357$ & $0.364$ & $0.367$ & $0.376$ & $0.405$ & $0.402$ & $\mathbf{0.405}$ & $0.369$ & $+0.15$ & $+0.029$ \\
 & Llama-3.1-8B & $0.353$ & $0.412$ & $0.444$ & $0.472$ & $0.498$ & $\mathbf{0.507}$ & $0.488$ & $0.467$ & $0.475$ & $+0.05$ & $+0.009$ \\
 & Qwen3-8B & $0.610$ & $0.639$ & $0.657$ & $\mathbf{0.668}$ & $0.661$ & $0.653$ & $0.626$ & $0.602$ & $0.576$ & $-0.05$ & $+0.006$ \\
 & Qwen3-30B-A3B & $0.347$ & $0.319$ & $0.311$ & $0.323$ & $0.345$ & $0.368$ & $0.394$ & $0.409$ & $\mathbf{0.412}$ & $+0.2$ & $+0.067$ \\
\addlinespace[2pt]
AGIEval & Llama-3.2-1B & $0.272$ & $0.274$ & $0.275$ & $0.276$ & $0.273$ & $\mathbf{0.277}$ & $0.270$ & $0.269$ & $0.267$ & $+0.05$ & $+0.004$ \\
 & Llama-3.2-3B & $0.294$ & $0.320$ & $0.336$ & $0.342$ & $\mathbf{0.351}$ & $0.349$ & $0.346$ & $0.336$ & $0.318$ & $0$ & $+0.000$ \\
 & Llama-3.1-8B & $0.334$ & $0.396$ & $0.411$ & $\mathbf{0.424}$ & $0.419$ & $0.416$ & $0.407$ & $0.382$ & $0.351$ & $-0.05$ & $+0.005$ \\
 & Qwen3-8B & $0.559$ & $0.578$ & $0.585$ & $\mathbf{0.588}$ & $0.585$ & $0.578$ & $0.570$ & $0.551$ & $0.513$ & $-0.05$ & $+0.003$ \\
 & Qwen3-30B-A3B & $0.637$ & $\mathbf{0.641}$ & $0.641$ & $0.636$ & $0.634$ & $0.625$ & $0.623$ & $0.605$ & $0.591$ & $-0.15$ & $+0.007$ \\
\addlinespace[2pt]
BBH & Llama-3.2-1B & $0.292$ & $0.297$ & $\mathbf{0.298}$ & $0.296$ & $0.296$ & $0.293$ & $0.290$ & $0.285$ & $0.269$ & $-0.1$ & $+0.002$ \\
 & Llama-3.2-3B & $0.308$ & $0.353$ & $0.377$ & $0.389$ & $\mathbf{0.399}$ & $0.390$ & $0.380$ & $0.356$ & $0.319$ & $0$ & $+0.000$ \\
 & Llama-3.1-8B & $0.386$ & $0.442$ & $0.450$ & $0.450$ & $\mathbf{0.451}$ & $0.439$ & $0.431$ & $0.407$ & $0.352$ & $0$ & $+0.000$ \\
 & Qwen3-8B & $0.509$ & $0.527$ & $\mathbf{0.534}$ & $0.527$ & $0.517$ & $0.493$ & $0.460$ & $0.421$ & $0.402$ & $-0.1$ & $+0.017$ \\
 & Qwen3-30B-A3B & $\mathbf{0.415}$ & $0.396$ & $0.377$ & $0.366$ & $0.364$ & $0.371$ & $0.370$ & $0.381$ & $0.380$ & $-0.2$ & $+0.051$ \\
\bottomrule
\end{tabular}%
}
\end{table*}

Table~\ref{tab:selected-subtask-gains} then opens the module averages and shows selected subtask-level cases. The table keeps the three judge-derived capability weights visible, so gains can be read jointly with task demand: some large shifts occur on deductive BBH subtasks, whereas many Qwen3 30B-A3B gains are concentrated in factual MMLU-Redux subjects. This selected view is intentionally diagnostic rather than exhaustive; Appendix~\ref{sec:appendix-full-subtasks} reports the complete 118-subtask table.

\begin{table*}[t]
\centering
\scriptsize
\setlength{\tabcolsep}{2.2pt}
\caption{Selected high-gain capability-labeled subtasks from the 118-subtask suite. Rows are the 20 subtasks with the largest positive gain under any of the three representative models shown here: Llama 3.1 8B, Qwen3 8B, and Qwen3 30B-A3B. The judge columns report the normalized three-capability demand weights used in the capability fit. For each model, Base is the $\alpha=0$ score, $\alpha^*$ is the best perturbation on the nine-point grid, and $\Delta_{\mathrm{frac}}$ is the gain over baseline in task-score units. The complete 118-subtask table is reported in Appendix~\ref{sec:appendix-full-subtasks}.}
\label{tab:selected-subtask-gains}
\resizebox{\textwidth}{!}{%
\begin{tabular}{llccc|rrr|rrr|rrr}
\toprule
Module & Subtask & FK & LU & DR & \multicolumn{3}{c|}{Llama 3.1 8B} & \multicolumn{3}{c|}{Qwen3 8B} & \multicolumn{3}{c}{Qwen3 30B-A3B} \\
\cmidrule(lr){6-8}\cmidrule(lr){9-11}\cmidrule(lr){12-14}
 & & & & & Base & $\alpha^*$ & $\Delta_{\mathrm{frac}}$ & Base & $\alpha^*$ & $\Delta_{\mathrm{frac}}$ & Base & $\alpha^*$ & $\Delta_{\mathrm{frac}}$ \\
\midrule
BBH & temporal\_sequences & $0.000$ & $0.147$ & $0.853$ & $0.816$ & $-0.1$ & $+0.120$ & $0.080$ & $-0.15$ & $+0.500$ & $0.024$ & $-0.2$ & $+0.500$ \\
MMLU-Redux & high\_school\_psychology & $0.888$ & $0.086$ & $0.026$ & $0.604$ & $+0.1$ & $+0.094$ & $0.844$ & $-0.05$ & $+0.042$ & $0.271$ & $+0.2$ & $+0.312$ \\
MMLU-Redux & logical\_fallacies & $0.790$ & $0.200$ & $0.010$ & $0.608$ & $+0.05$ & $+0.081$ & $0.811$ & $-0.1$ & $+0.027$ & $0.324$ & $-0.2$ & $+0.297$ \\
MMLU-Redux & high\_school\_us\_history & $0.630$ & $0.291$ & $0.079$ & $0.580$ & $0$ & $+0.000$ & $0.740$ & $+0.05$ & $+0.030$ & $0.360$ & $+0.2$ & $+0.260$ \\
MMLU-Redux & high\_school\_computer\_science & $0.647$ & $0.088$ & $0.265$ & $0.440$ & $+0.1$ & $+0.040$ & $0.770$ & $+0.05$ & $+0.050$ & $0.260$ & $+0.2$ & $+0.220$ \\
BBH & ruin\_names & $0.185$ & $0.815$ & $0.000$ & $0.748$ & $0$ & $+0.000$ & $0.508$ & $-0.2$ & $+0.068$ & $0.332$ & $-0.2$ & $+0.216$ \\
BBH & navigate & $0.000$ & $0.114$ & $0.886$ & $0.332$ & $-0.15$ & $+0.216$ & $0.704$ & $+0.05$ & $+0.008$ & $0.628$ & $+0.15$ & $+0.024$ \\
MMLU-Redux & high\_school\_government\_and\_politics & $0.872$ & $0.092$ & $0.036$ & $0.510$ & $+0.05$ & $+0.120$ & $0.810$ & $+0.05$ & $+0.040$ & $0.290$ & $+0.2$ & $+0.210$ \\
MMLU-Redux & college\_biology & $0.879$ & $0.061$ & $0.059$ & $0.582$ & $0$ & $+0.000$ & $0.786$ & $0$ & $+0.000$ & $0.367$ & $+0.15$ & $+0.204$ \\
MMLU-Redux & high\_school\_world\_history & $0.580$ & $0.358$ & $0.062$ & $0.556$ & $-0.05$ & $+0.040$ & $0.737$ & $-0.1$ & $+0.081$ & $0.293$ & $+0.2$ & $+0.202$ \\
MMLU-Redux & astronomy & $0.924$ & $0.039$ & $0.037$ & $0.564$ & $+0.05$ & $+0.021$ & $0.777$ & $+0.05$ & $+0.032$ & $0.362$ & $+0.2$ & $+0.202$ \\
MMLU-Redux & anatomy & $0.955$ & $0.034$ & $0.010$ & $0.525$ & $-0.05$ & $+0.061$ & $0.626$ & $-0.05$ & $+0.071$ & $0.263$ & $+0.1$ & $+0.182$ \\
BBH & reasoning\_about\_colored\_objects & $0.030$ & $0.507$ & $0.463$ & $0.588$ & $+0.1$ & $+0.012$ & $0.884$ & $-0.1$ & $+0.056$ & $0.548$ & $-0.2$ & $+0.168$ \\
MMLU-Redux & college\_computer\_science & $0.583$ & $0.096$ & $0.321$ & $0.330$ & $-0.1$ & $+0.062$ & $0.598$ & $0$ & $+0.000$ & $0.299$ & $+0.2$ & $+0.165$ \\
MMLU-Redux & high\_school\_european\_history & $0.566$ & $0.341$ & $0.092$ & $0.473$ & $-0.05$ & $+0.055$ & $0.769$ & $0$ & $+0.000$ & $0.352$ & $+0.2$ & $+0.165$ \\
MMLU-Redux & high\_school\_microeconomics & $0.707$ & $0.105$ & $0.188$ & $0.480$ & $-0.05$ & $+0.051$ & $0.837$ & $0$ & $+0.000$ & $0.327$ & $+0.1$ & $+0.163$ \\
MMLU-Redux & sociology & $0.930$ & $0.066$ & $0.004$ & $0.673$ & $0$ & $+0.000$ & $0.765$ & $-0.1$ & $+0.031$ & $0.388$ & $+0.1$ & $+0.153$ \\
MMLU-Redux & college\_mathematics & $0.299$ & $0.117$ & $0.584$ & $0.222$ & $+0.05$ & $+0.152$ & $0.384$ & $-0.15$ & $+0.061$ & $0.222$ & $-0.2$ & $+0.051$ \\
MMLU-Redux & human\_aging & $0.947$ & $0.045$ & $0.008$ & $0.563$ & $+0.1$ & $+0.023$ & $0.690$ & $+0.05$ & $+0.023$ & $0.356$ & $+0.1$ & $+0.149$ \\
MMLU-Redux & moral\_scenarios & $0.250$ & $0.750$ & $0.000$ & $0.237$ & $+0.05$ & $+0.103$ & $0.299$ & $-0.15$ & $+0.144$ & $0.216$ & $+0.15$ & $+0.052$ \\
\bottomrule
\end{tabular}%
}
\end{table*}

\begin{figure*}[t]
\centering
\begin{minipage}[t]{0.32\textwidth}
\centering
\includegraphics[width=\linewidth]{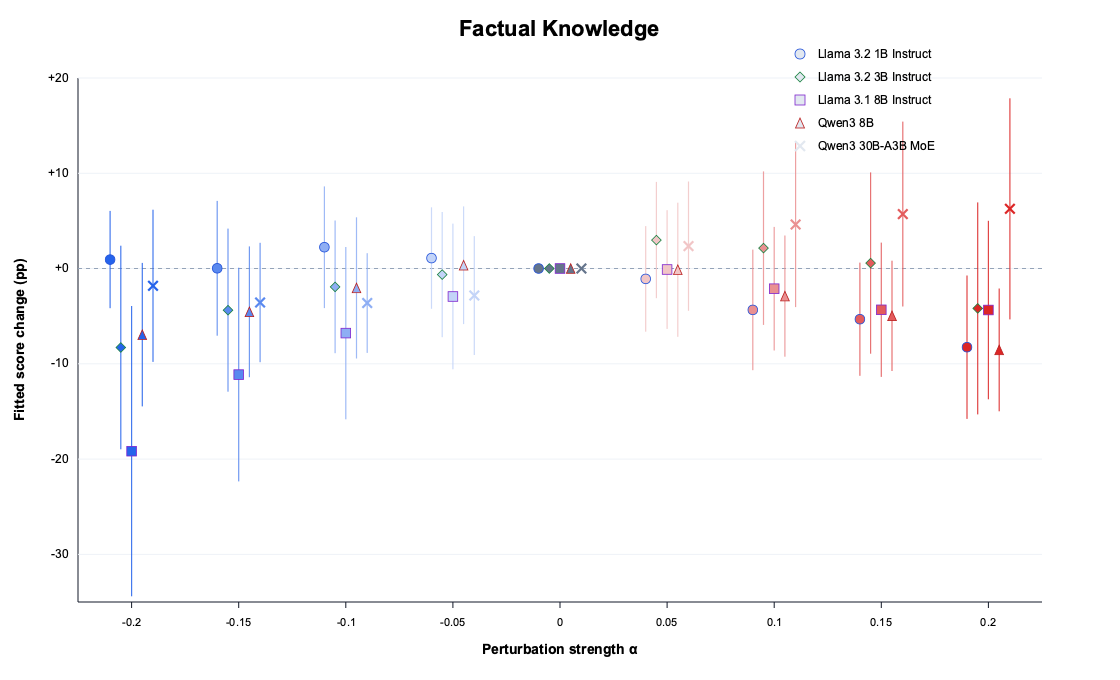}

\vspace{2pt}
\footnotesize\textbf{(a)} Factual Knowledge.
\end{minipage}\hfill
\begin{minipage}[t]{0.32\textwidth}
\centering
\includegraphics[width=\linewidth]{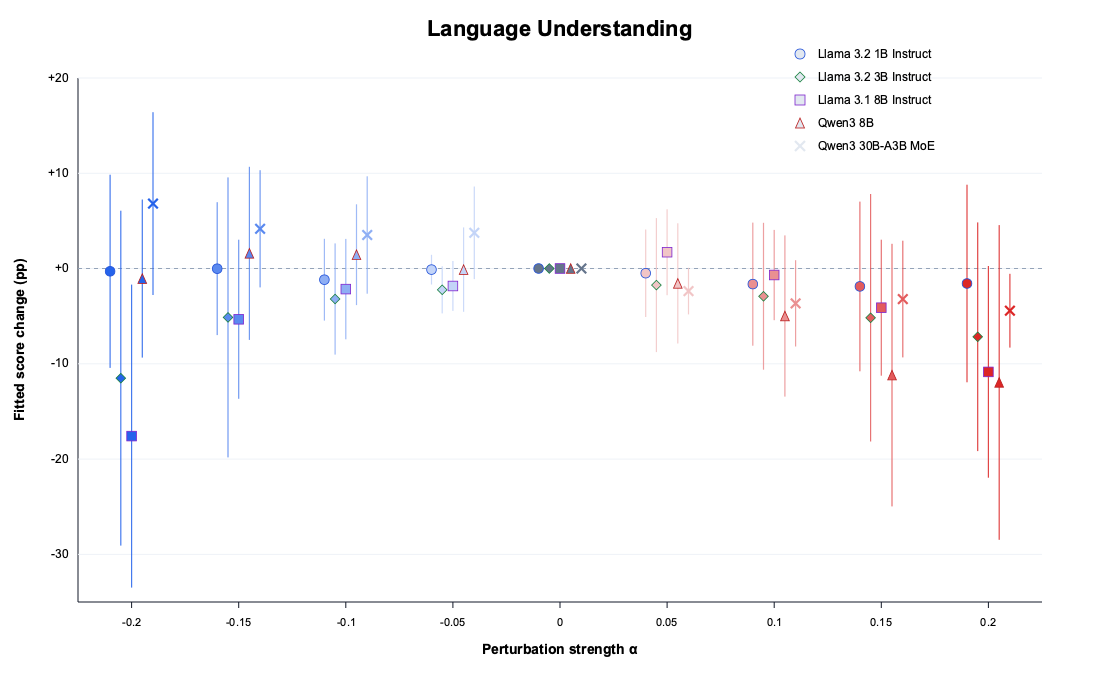}

\vspace{2pt}
\footnotesize\textbf{(b)} Language Understanding.
\end{minipage}\hfill
\begin{minipage}[t]{0.32\textwidth}
\centering
\includegraphics[width=\linewidth]{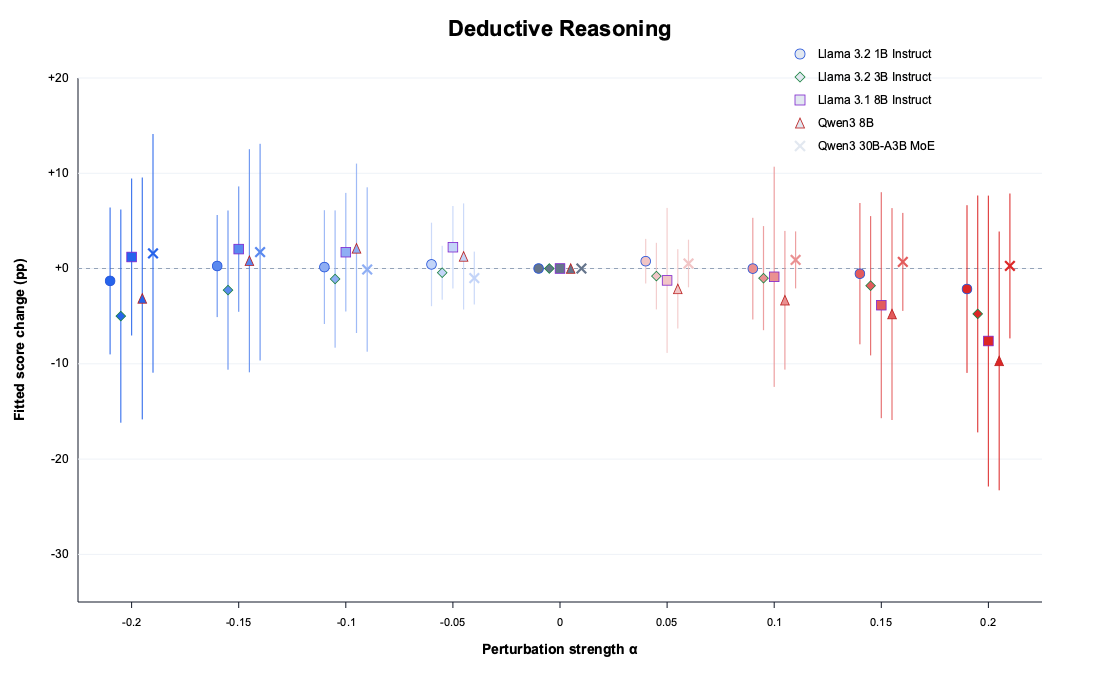}

\vspace{2pt}
\footnotesize\textbf{(c)} Deductive Reasoning.
\end{minipage}
\caption{Capability-level perturbation responses fitted from the 118 subtask score changes. Each panel shows one retained capability dimension, with points denoting the maximum-likelihood fitted score change relative to the unperturbed model at each $\alpha$ value. Error bars show one fitted residual standard deviation for that capability and perturbation strength; they summarize subtask heterogeneity and are not formal 95\% confidence intervals.}
\label{fig:eval6-three-capability-errorbars}
\end{figure*}

We then ask whether the perturbation acts like generic noise or like a structured reweighting of latent capabilities. Figure~\ref{fig:eval6-three-capability-errorbars} plots fitted score deltas for the retained \emph{Factual Knowledge}, \emph{Language Understanding}, and \emph{Deductive Reasoning} axes. The dominant pattern is asymmetric rather than monotone: moderate negative $\alpha$ often preserves or slightly improves factual or language-heavy behavior, whereas large positive $\alpha$ more consistently suppresses all three axes, especially language and deductive components.

Model family and scale matter as much as perturbation sign. The smaller dense Llama models move more sharply as $\alpha$ departs from zero, while the larger Qwen3 30B-A3B MoE is comparatively stable at the union level. That architecture dependence suggests that the transform is not simply injecting random damage; it is interacting with pre-existing spectral organization in a non-uniform way.

The fitted capability picture matches the raw module-level evidence rather than replacing it. Three regularities are especially important. First, the high-positive interval drives all three capability responses downward, with the clearest contractions on language and deductive axes. Second, negative $\alpha$ is more heterogeneous and sometimes produces factual or language recovery rather than monotone degradation. Third, union-level averages conceal module-specific stress: Appendix~\ref{sec:appendix-module-wise} and Table~\ref{tab:module-sensitivity-summary} show that the Llama family is most sensitive in \texttt{MMLU-Pro}, whereas Qwen models are at least as sensitive to \texttt{BBH}. Table~\ref{tab:module-total-all-alpha} further shows that each module can prefer a different $\alpha$, reinforcing that the perturbation reorganizes capability balance rather than moving a single performance scalar.

\section{Trait Responses Track Capability Reweighting}
\label{sec:trait-responses}

\begin{figure*}[t]
\centering
\includegraphics[width=\textwidth]{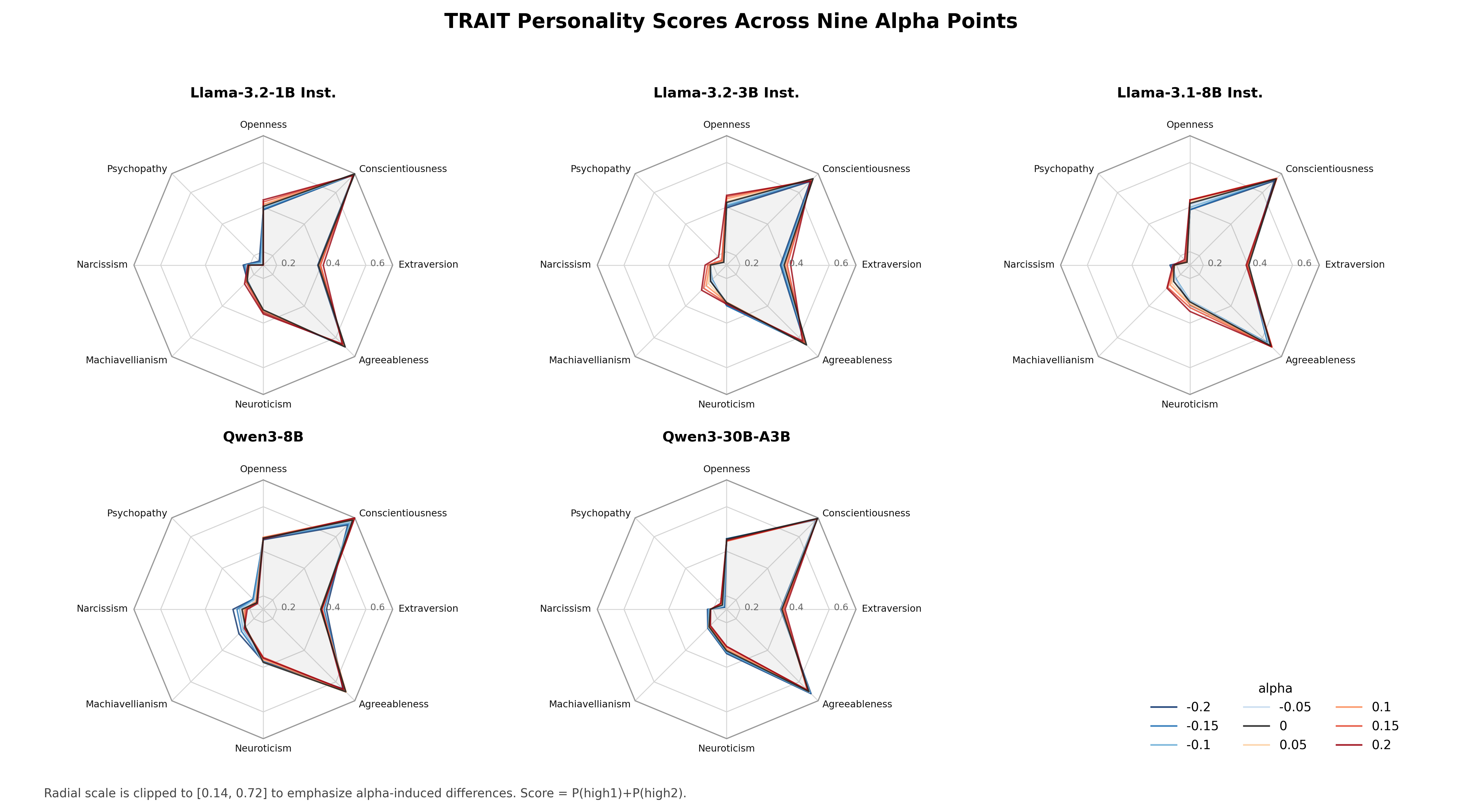}
\caption{TRAIT personality profiles across the full nine-point $\alpha$ grid for the five evaluated models. Each radar chart overlays all tested perturbation strengths for a single model. The radial scale is clipped to emphasize perturbation-induced differences rather than absolute trait ordering.}
\label{fig:trait-radar-grid}
\end{figure*}

The personality-style probes show that $\alpha$ perturbations do not only move benchmark accuracy. Figure~\ref{fig:trait-radar-grid} overlays the full nine-point $\alpha$ sweep for each model in the TRAIT space. The absolute ordering of traits is fairly stable---Agreeableness and Conscientiousness remain high while Narcissism and Psychopathy stay low---but the radar profiles rotate systematically with the perturbation, especially along Openness, Agreeableness, Machiavellianism, and Neuroticism. Because the radial range is clipped, the figure emphasizes within-model perturbation effects rather than raw between-trait level differences.

\paragraph{TRAIT scoring protocol.}
TRAIT uses the full available split for each of the eight published dimensions: Openness, Conscientiousness, Extraversion, Agreeableness, Neuroticism, Machiavellianism, Narcissism, and Psychopathy. For every item, we score the four provided continuations (two high-trait and two low-trait responses) under the model's conditional log-likelihood given the prompt \texttt{Question: \{q\}} followed by a newline and \texttt{Response: }. The four candidate responses are shuffled deterministically with seed 1234 keyed by split and item index. The item-level trait score is the softmax probability mass assigned to the two high-trait responses, and the reported trait score is the average of that quantity over all items in the split. Because this probe is continuation-likelihood based rather than free-form generation based, no answer-extraction heuristic is involved.

\begin{figure*}[t]
\centering
\begin{minipage}[t]{0.48\textwidth}
\centering
\includegraphics[width=\linewidth]{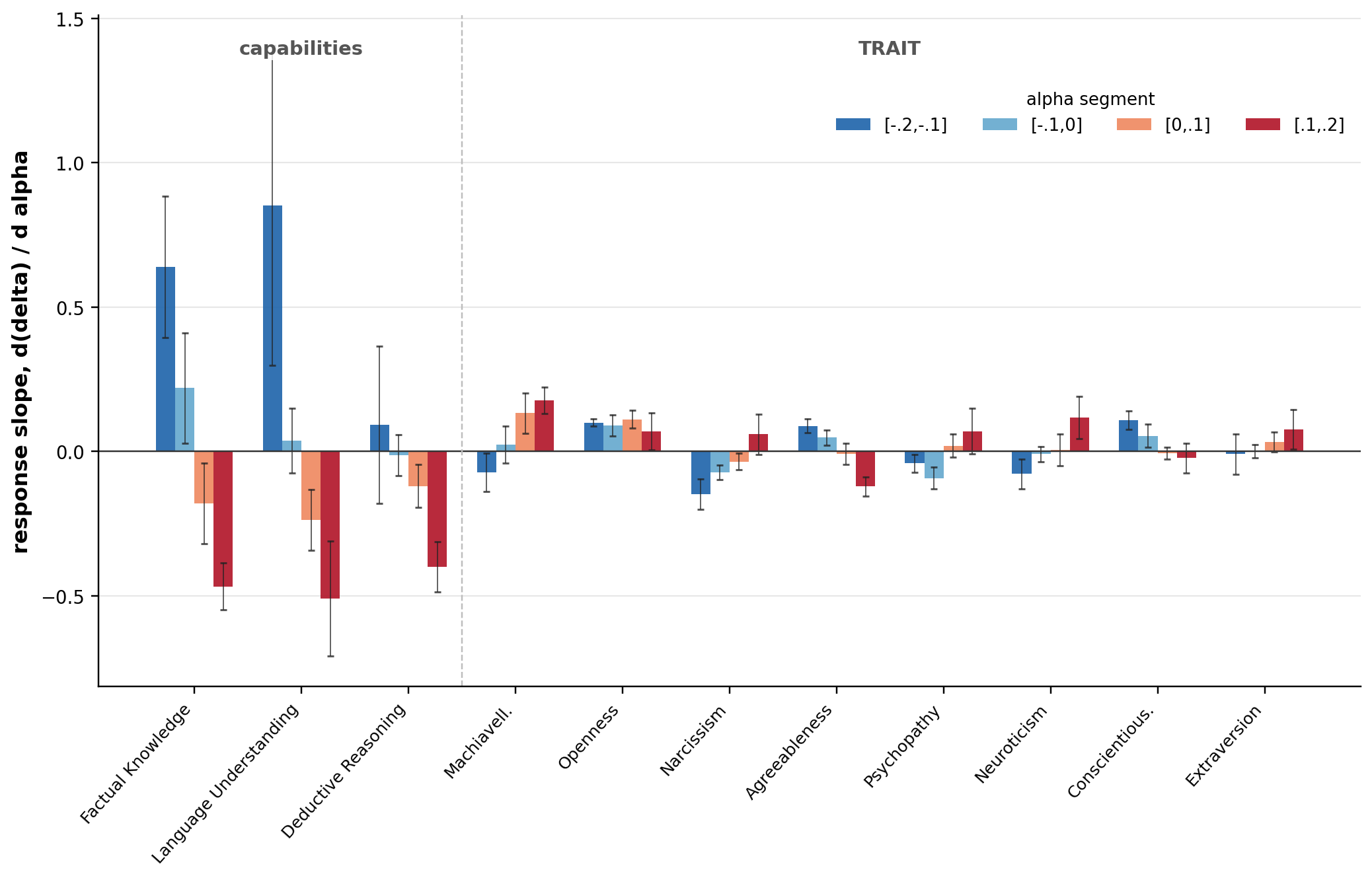}

\vspace{2pt}
\footnotesize\textbf{(a)} Local response slopes across four $\alpha$ intervals.
\end{minipage}\hfill
\begin{minipage}[t]{0.48\textwidth}
\centering
\includegraphics[width=\linewidth]{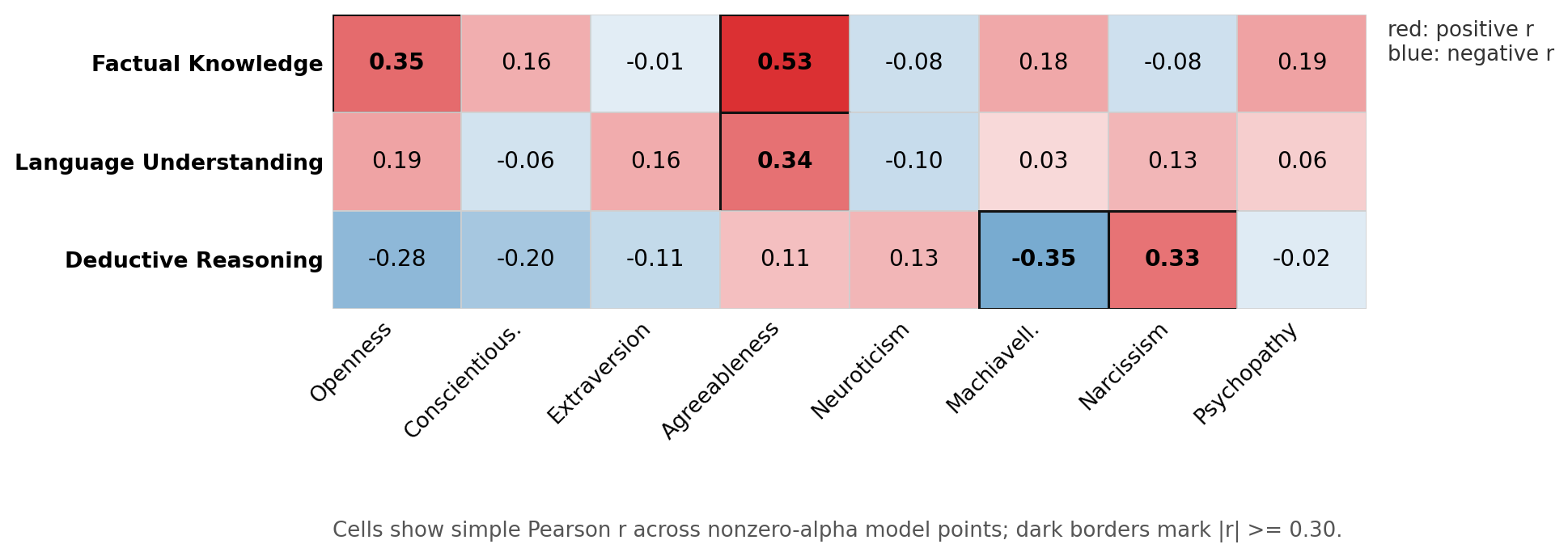}

\vspace{2pt}
\footnotesize\textbf{(b)} Pearson correlations between capability and trait responses.
\end{minipage}

\vspace{8pt}
\begin{minipage}[t]{\textwidth}
\centering
\includegraphics[width=0.94\linewidth]{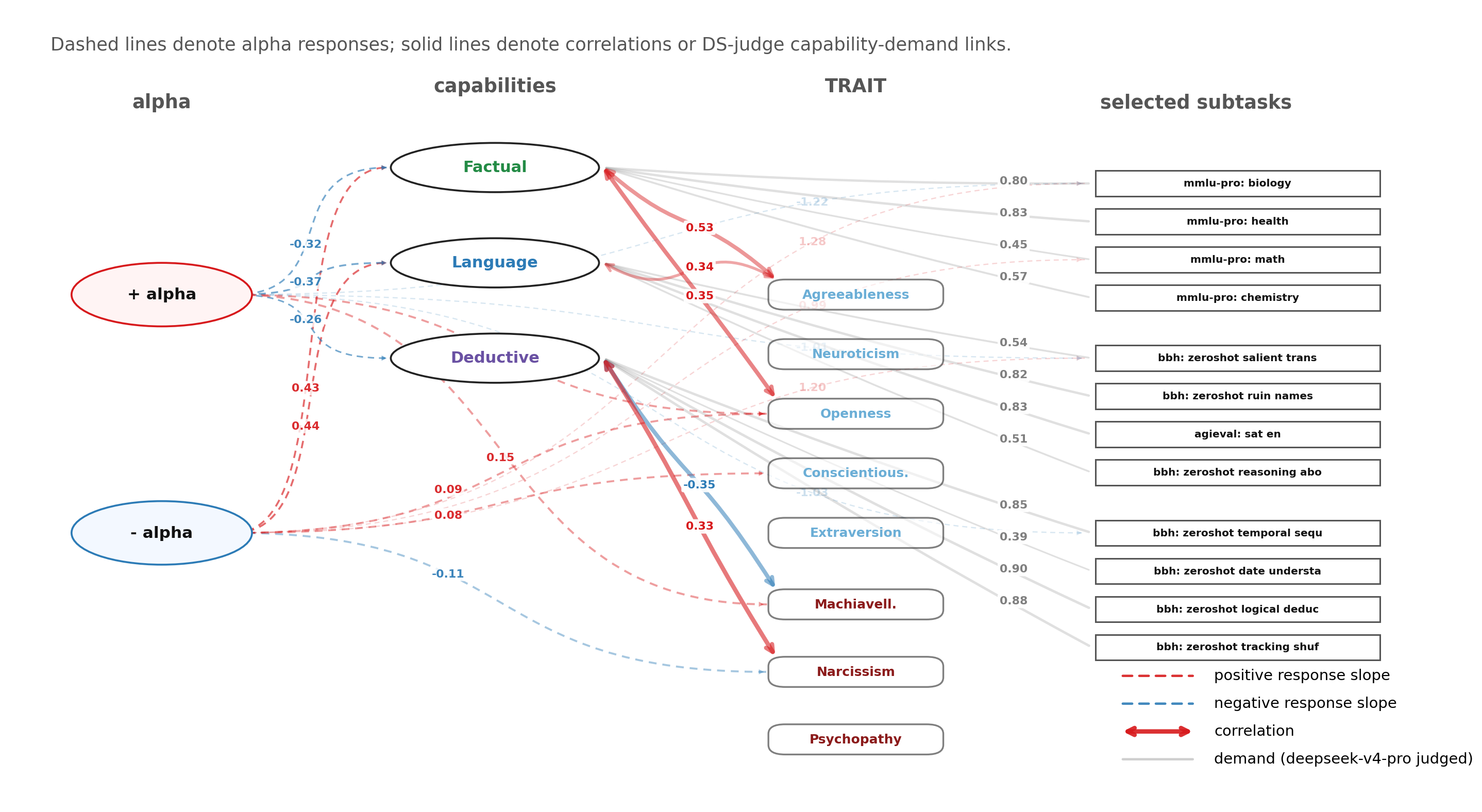}

\vspace{2pt}
\footnotesize\textbf{(c)} Association network linking $\alpha$ direction, latent capabilities, TRAIT dimensions, and selected subtasks.
\end{minipage}
\caption{Alpha perturbation responses link latent capability shifts, TRAIT shifts, and task-level demands. Panel (a) shows mean local response slopes across four dense instruct models, with error bars denoting SEM across models. Panel (b) reports descriptive Pearson correlations across nonzero-$\alpha$ points from the four dense models ($n=32$; Qwen3 30B-A3B MoE excluded) and highlights cells with $|r|\ge 0.30$; because points from the same model share a baseline, these cells are exploratory associations rather than independent-sample significance claims. Panel (c) combines directional response slopes, capability--TRAIT correlations, and DeepSeek-V4-Pro capability-demand links for selected subtasks.}
\label{fig:trait-capability-links}
\end{figure*}

Figure~\ref{fig:trait-capability-links} then ties those trait movements back to the capability fits. The strongest local slope effects occur at the ends of the $\alpha$ range: factual knowledge increases on the more negative interval, while large positive $\alpha$ suppresses factual, language, and deductive responses. On the correlation side, we report descriptive Pearson correlations over nonzero-$\alpha$ points from the four dense models ($n=32$, excluding Qwen3 30B-A3B MoE): Factual Knowledge aligns positively with Agreeableness ($r=0.53$) and Openness ($r=0.35$), while Deductive Reasoning is negatively associated with Machiavellianism ($r=-0.35$) and positively associated with Narcissism ($r=0.33$). These correlations should be interpreted as exploratory associations across perturbation settings rather than formal significance tests, because the eight $\alpha$ points from the same model share a baseline and model-specific response curve. The network view therefore illustrates, rather than proves, how selective task gains can coexist with global degradation: different $\alpha$ directions pull capabilities, traits, and subtask families along partially overlapping but non-identical axes.

Importantly, we do not interpret the TRAIT scores as stable human-like personalities. In this paper they function as compact behavioral summaries that can be compared across perturbation settings. Their value is that they move in coordinated ways with latent capability responses: the same high-positive $\alpha$ interval that suppresses language and deductive responses also tends to lower Agreeableness and raise Machiavellianism. The joint capability--trait view therefore supports the interpretation that spectral perturbations reshape bundles of behaviors rather than isolated benchmark scores.

\section{Interpretation, Limitations, and Outlook}
\label{sec:discussion}

Across capability fits, auxiliary tasks, and trait probes, the common effect of spectral smoothing and spiking is reconfiguration rather than uniform collapse. Positive $\alpha$ often acts as a high-variance stressor that selectively harms language and deductive behavior, while moderate negative $\alpha$ can occasionally recover factual or task-specific performance. The layer-wise evidence in Appendix~\ref{sec:appendix-layerwise} suggests a mechanistic reason: one nominal $\alpha$ induces non-uniform realized updates because layers begin from different degrees of spectral inequality.

\paragraph{Why selective gains occur.}
If every layer responded equally to the same $\alpha$, one would expect a smoother tradeoff between average quality and perturbation strength. Instead, the appendix diagnostics show that layers with already concentrated spectra receive larger realized changes, and the most affected layers differ by projection type and model family. This makes the intervention behave more like a non-uniform reweighting of internal computations than like a scalar noise injection.

\paragraph{Limitations.}
The study covers five open models, fixed $\alpha$ grids, and benchmark suites with different scales and evaluation noise. Some measurements rely on external judges, and the retained three-dimensional capability basis is benchmark-specific rather than a universal ontology of language-model competence. We therefore treat selective gains as diagnostic evidence about capability organization, not as a general-purpose recipe for improving models.

\paragraph{Outlook.}
Even with those limits, the intervention is attractive because it is post-hoc, architecture-aware, and easy to audit. The results motivate future variants such as layer-normalized $\alpha$, projection-specific controls for \texttt{up\_proj} and \texttt{down\_proj}, or expert-wise safeguards for MoE models. More broadly, they suggest that weight-space spectral statistics are a useful handle for probing how different capabilities are bundled together in large language models.

\section{Conclusion}
\label{sec:conclusion}

We introduced spectral smoothing and spiking of transformer MLP singular values as a diagnostic intervention and evaluated it across capability-labeled subtasks, auxiliary tasks, judge-scored truthfulness/hallucination/safety tasks, and TRAIT personality probes. The main result is evidence of a non-uniform reconfiguration of behavior rather than uniform degradation: large positive perturbations more reliably suppress language and deductive responses, moderate negative perturbations can sometimes recover factual or task-specific performance on the evaluated grids, and the magnitude of these effects depends on model family and layerwise spectral inequality. These findings support the view that singular-value spectra encode part of the organization of model capabilities and provide a useful handle for probing intervention-induced tradeoffs without retraining.

\clearpage

\appendix

\section{Justification for the Three-Dimensional Capability Basis}
\label{sec:appendix-capability-basis}

The main paper analyzes subtask response profiles in a three-dimensional basis consisting of \emph{Factual Knowledge}, \emph{Language Understanding}, and \emph{Deductive Reasoning}. This appendix explains why that reduced basis is a useful low-dimensional summary of the current benchmark geometry. We use the original seven-way judge mixtures to identify the exposed benchmark-wide axes, and then use a separate constrained three-way judge pass to obtain the normalized weights used in the capability-response fit.

\begin{figure*}[t]
\centering
\begin{minipage}[t]{0.49\textwidth}
\centering
\includegraphics[width=\linewidth,height=0.34\textheight,keepaspectratio]{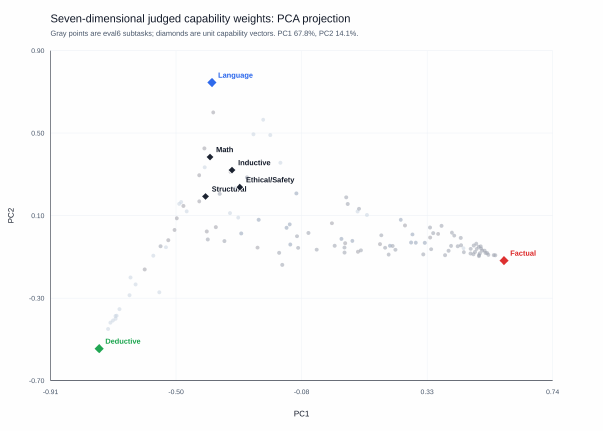}

\vspace{2pt}
\footnotesize\textbf{(a)} PCA projection of the 118 benchmark-subtask mixtures (gray points) together with the seven unit-dimension markers.
\end{minipage}\hfill
\begin{minipage}[t]{0.49\textwidth}
\centering
\includegraphics[width=\linewidth,height=0.34\textheight,keepaspectratio]{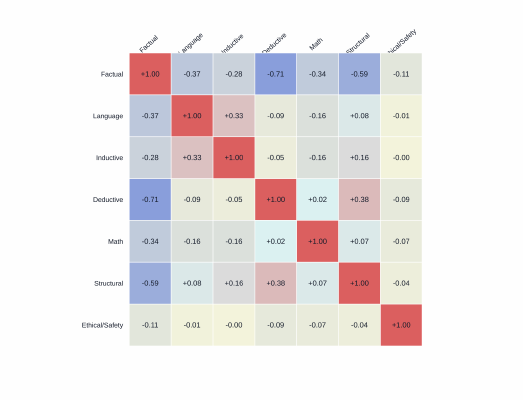}

\vspace{2pt}
\footnotesize\textbf{(b)} Pairwise correlations among the seven original judge dimensions.
\end{minipage}
\caption{Two complementary views of the seven-dimension capability geometry. The PCA view shows that the most exposed vertices are \emph{Factual Knowledge}, \emph{Language Understanding}, and \emph{Deductive Reasoning}. The correlation matrix shows that several remaining dimensions behave as composites or complements rather than as independent benchmark-wide extremes.}
\label{fig:seven-dim-capability-basis}
\end{figure*}

\paragraph{PCA evidence.}
The first two principal components explain 67.8\% and 14.1\% of the variance, respectively, for 81.9\% cumulative variance in the seven-dimensional judge mixtures. PC1 is dominated by a contrast between \emph{Factual Knowledge} ($+0.8579$) and \emph{Deductive Reasoning} ($-0.4739$), while PC2 mainly separates \emph{Language Understanding} ($+0.5717$) from \emph{Deductive Reasoning} ($-0.7195$). In the PCA plane, these three retained dimensions define the outer vertices of the observed task cloud, whereas the remaining four dimensions lie closer to the interior or along directions between those vertices.

\paragraph{Core-triangle geometry.}
Using the three exposed unit vectors as vertices of a triangle in the PCA plane, all 118 benchmark-subtask points fall inside that triangle. The omitted dimensions are also interior when expressed in barycentric coordinates over the retained basis: \emph{Inductive Reasoning} $=(0.156, 0.620, 0.225)$, \emph{Mathematical Computation} $=(0.081, 0.693, 0.226)$, \emph{Structural Analysis} $=(0.114, 0.534, 0.352)$, and \emph{Ethical \& Safety Judgment} $=(0.197, 0.542, 0.262)$ over $(\text{Factual}, \text{Language}, \text{Deductive})$. This indicates that, within this benchmark collection, these dimensions behave as mixtures or localized specializations rather than as additional benchmark-wide extremes.

\paragraph{Correlation and reconstruction checks.}
Because the judge weights live on a simplex, the correlation matrix reflects both semantic co-occurrence and competition for probability mass, but it still reveals redundancy structure. The largest absolute pairwise correlations are between \emph{Factual Knowledge} and \emph{Deductive Reasoning} ($-0.7089$), between \emph{Factual Knowledge} and \emph{Structural Analysis} ($-0.5893$), and between \emph{Deductive Reasoning} and \emph{Structural Analysis} ($+0.3826$). Regressing each original dimension on the retained three perfectly reconstructs the retained axes themselves and partially reconstructs several omitted ones: \emph{Mathematical Computation} reaches $R^2 = 0.5338$, \emph{Structural Analysis} reaches $R^2 = 0.3934$, and \emph{Inductive Reasoning} reaches $R^2 = 0.2046$, whereas \emph{Ethical \& Safety Judgment} remains mostly residual with $R^2 = 0.1194$. Low $R^2$ here does not imply that a dimension is conceptually unimportant; it only means that, in this benchmark collection, it does not appear as a broad independent axis comparable to the three PCA-exposed vertices.

\paragraph{Implication for the main paper.}
The three-dimensional basis used for the main alpha-response analysis is therefore not a replacement for the conceptual seven-dimension rubric. It is a data-driven low-dimensional basis for the current 118-subtask geometry: it identifies the dominant outer structure of judged task demands, after which the final FK/LU/DR weights are obtained by the separate constrained three-way labeling pass used in the fits.

\section{Module-wise Three-Dimensional Capability Response Analysis}
\label{sec:appendix-module-wise}

This appendix summarizes independent capability fits across four benchmark modules and their union. The retained capability basis contains three dimensions: \emph{Factual Knowledge}, \emph{Language Understanding}, and \emph{Deductive Reasoning}, with subtask weights taken from the constrained three-way judge pass described in Section~\ref{sec:capability-responses}. For each model, module, $\alpha$, and dimension, the plotted value is the maximum-likelihood capability-level score delta relative to $\alpha=0$, and the error bars show the fitted residual standard deviation for that coordinate. Thus the bars quantify how much subtask-to-subtask variation remains after projection onto the three retained capabilities; they are not 95\% confidence intervals on the mean. The five columns are \texttt{MMLU-Pro}, \texttt{MMLU-Redux}, \texttt{AGIEval}, \texttt{BBH}, and \texttt{All}.

The corresponding raw module means are reported in the main paper in Table~\ref{tab:module-total-all-alpha}; this appendix focuses on the independent module-wise capability fits and residual diagnostics.

\begin{figure*}[t]
\centering
\includegraphics[width=\textwidth]{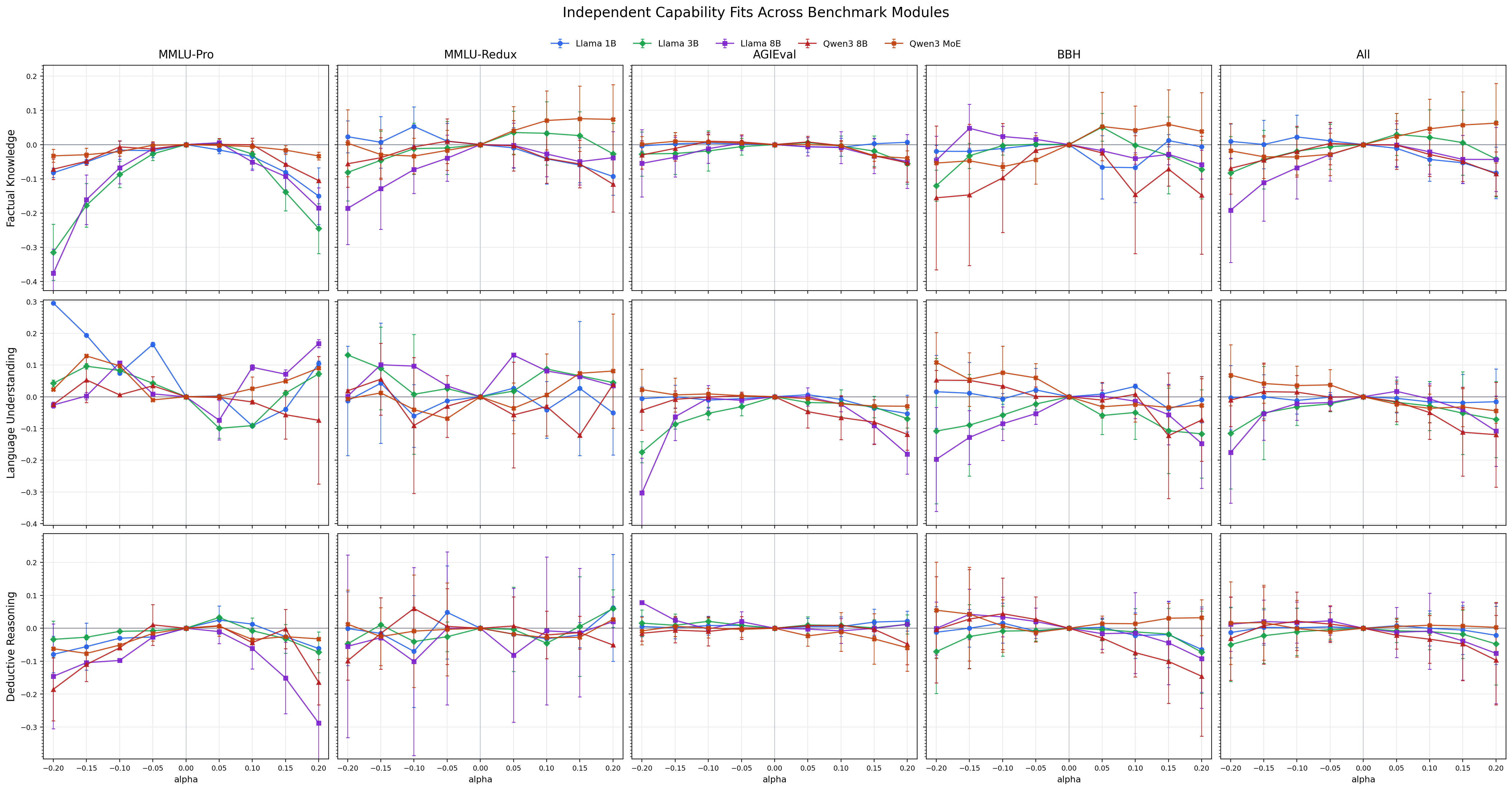}
\caption{Independent three-dimensional capability fits across benchmark modules and their union. Each point shows the fitted capability-level score delta relative to $\alpha=0$ for a fixed model, module, perturbation strength, and retained capability axis. Error bars are fitted one-standard-deviation residual scales, not formal 95\% confidence intervals; Table~\ref{tab:capability-fit-diagnostics} reports residual diagnostics for the same fits.}
\label{fig:module-wise-capability-fits}
\end{figure*}

\begin{figure}[t]
\centering
\includegraphics[width=0.8\linewidth]{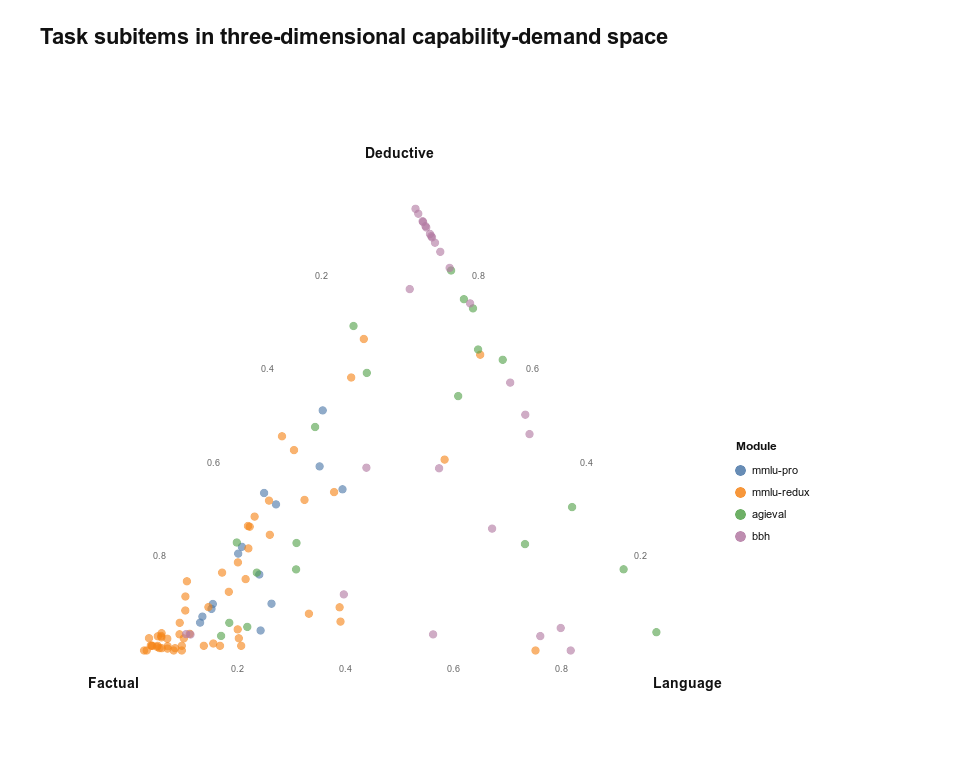}
\caption{Task subitems in the three-dimensional capability-demand simplex. Each point denotes a benchmark subitem, colored by evaluation module. The three vertices correspond to factual knowledge, language understanding, and deductive reasoning demand weights judged by DeepSeek-V4-Pro.}
\label{fig:task-capability-simplex}
\end{figure}

\paragraph{Coverage.}
All five models have complete coverage in all four modules: 14 \texttt{MMLU-Pro} subtasks, 57 \texttt{MMLU-Redux} subtasks, 20 \texttt{AGIEval} subtasks, and 27 \texttt{BBH} subtasks, for 118 subtasks in the union fit. The previously missing \emph{Llama 3.1 8B / BBH / $\alpha=-0.1$} and \emph{Qwen3 30B-A3B / AGIEval / $\alpha=-0.2$} points are included here.

\begin{table}[t]
\centering
\small
\begin{tabular}{lrrrr}
\toprule
Scope & Subtasks & Observations & RMSE & $R^2$ \\
\midrule
\texttt{MMLU-Pro} & 14 & 560 & 0.030 & 0.824 \\
\texttt{MMLU-Redux} & 57 & 2280 & 0.064 & 0.313 \\
\texttt{AGIEval} & 20 & 800 & 0.028 & 0.519 \\
\texttt{BBH} & 27 & 1080 & 0.073 & 0.250 \\
All & 118 & 4720 & 0.064 & 0.261 \\
\bottomrule
\end{tabular}
\caption{Residual diagnostics for the three-dimensional capability fits, pooled over five models and the eight nonzero $\alpha$ values. RMSE is computed on subtask score deltas after projecting through the fitted capability coordinates. The moderate union-level $R^2$ is expected because the three retained axes summarize broad capability responses rather than every task-specific idiosyncrasy; therefore conclusions are emphasized only when they also agree with raw module means and bootstrap/error-bar scale.}
\label{tab:capability-fit-diagnostics}
\end{table}

\paragraph{Main patterns.}
The strongest module-local response appears in \texttt{MMLU-Pro}. Across models, \texttt{MMLU-Pro} shows the largest average absolute capability shifts, especially for Llama 3.1 8B and Llama 3.2 3B, making it the most sensitive component of the current suite to singular-value perturbation. \texttt{AGIEval} is comparatively stable for Llama 3.2 1B and Qwen3 30B-A3B MoE, but not for Llama 3.1 8B, which shows large negative \emph{Language Understanding} shifts at both $\alpha=-0.2$ and $\alpha=+0.2$. \texttt{BBH} is the most model-dependent module: Qwen3 8B shows strong positive-$\alpha$ degradation, especially in factual and deductive directions, whereas Qwen3 30B-A3B MoE is more balanced and less consistently negative.

\paragraph{Model-wise interpretation.}
The global response of Llama 3.2 1B Instruct is mild, with the largest local movements appearing in \texttt{MMLU-Pro}. Llama 3.2 3B Instruct is more sensitive, especially at $\alpha=-0.2$, where all three union-fit dimensions become negative and the largest shift is in \emph{Language Understanding}. Llama 3.1 8B Instruct is the most sensitive dense model: at $\alpha=-0.2$, its union fit reaches approximately $-0.192$ in \emph{Factual Knowledge} and $-0.176$ in \emph{Language Understanding}, with additional large degradations in \texttt{AGIEval} and \texttt{BBH}. Qwen3 8B is more asymmetric toward positive $\alpha$, with all three union dimensions negative at $\alpha=+0.2$. Qwen3 30B-A3B MoE has the smallest union-level response; its shifts are weaker and less consistently negative, suggesting that expert routing or model capacity absorbs part of the perturbation even though module-level fits still show localized movements.

\begin{table}[t]
\centering
\small
\begin{tabular}{lll}
\toprule
Model & Strongest local module & Avg. abs. shift \\
\midrule
Llama 3.2 1B Instruct & \texttt{MMLU-Pro} & 0.072 \\
Llama 3.2 3B Instruct & \texttt{MMLU-Pro} & 0.074 \\
Llama 3.1 8B Instruct & \texttt{MMLU-Pro} & 0.099 \\
Qwen3 8B & \texttt{BBH} & 0.067 \\
Qwen3 30B-A3B MoE & \texttt{BBH} & 0.043 \\
\bottomrule
\end{tabular}
\caption{Average absolute nonzero fitted shifts by module. The dominant perturbation-sensitive module is not universal: the Llama family is most stressed by \texttt{MMLU-Pro}, whereas the Qwen models are at least as sensitive to \texttt{BBH}.}
\label{tab:module-sensitivity-summary}
\end{table}

\paragraph{Interpretation.}
The three-dimensional basis is useful because it exposes where a global spectral perturbation becomes a module-specific capability shift. The same $\alpha$ value can manifest as factual degradation in \texttt{MMLU-Pro}, language degradation in \texttt{AGIEval}, or deductive degradation in \texttt{BBH}, depending on model family and scale. The union fit remains useful for an aggregate statement, but the module-wise fits are necessary for interpretation because several global responses are produced by localized module stress rather than by smooth degradation across all task types.

\section{Full Subtask-Level Best-Alpha Table}
\label{sec:appendix-full-subtasks}

Table~\ref{tab:eval6-best-alpha-subtasks} reports the complete 118-subtask version of the selected subtask summary in Table~\ref{tab:selected-subtask-gains}. It includes all four benchmark modules, the three retained judge-derived capability weights, and the baseline, best perturbation, and gain for all five evaluated models.

\definecolor{AlphaPos}{HTML}{DC2626}
\definecolor{AlphaNeg}{HTML}{2563EB}
\definecolor{AlphaZero}{HTML}{9CA3AF}
\definecolor{CapZero}{HTML}{F3F4F6}
\definecolor{FKColor}{HTML}{22C55E}
\definecolor{LUColor}{HTML}{3B82F6}
\definecolor{DRColor}{HTML}{A855F7}
\begin{landscape}
\tiny
\setlength{\tabcolsep}{1.45pt}
\setlength{\extrarowheight}{0.35pt}
\renewcommand{\arraystretch}{0.88}
\begin{longtable}{|S|rrr|rrr|rrr|rrr|rrr|rrr|}
\caption{Best perturbation point for each capability-labeled benchmark subtask and model. DS judge columns report the normalized three-capability demand weights: factual knowledge (FK; green), language understanding (LU; blue), and deductive reasoning (DR; purple), with cell backgrounds interpolating from white-gray at 0 to the pure capability color at 1. For each model, columns report the alpha-0 baseline score, the best alpha, and the fractional gain over baseline. Best-alpha values use blue text for negative alpha, red text for positive alpha, and gray text for alpha zero; saturation indicates $|\alpha|$. Fractional gain values above 0.05 are shown in \textbf{bold}; values above 0.1 are shown in \textbf{bold with an underline}.}\label{tab:eval6-best-alpha-subtasks}\\
\hline
\multirow{2}{*}{Subtask} & \multicolumn{3}{c|}{DS judge} & \multicolumn{3}{c|}{Llama 1B} & \multicolumn{3}{c|}{Llama 3B} & \multicolumn{3}{c|}{Llama 8B} & \multicolumn{3}{c|}{Qwen 8B} & \multicolumn{3}{c|}{Qwen 30B-A3B} \\
 & \textcolor{FKColor!70!black}{FK} & \textcolor{LUColor!85!black}{LU} & \textcolor{DRColor!85!black}{DR} & Base & $\alpha^*$ & $\Delta_{\mathrm{frac}}$ & Base & $\alpha^*$ & $\Delta_{\mathrm{frac}}$ & Base & $\alpha^*$ & $\Delta_{\mathrm{frac}}$ & Base & $\alpha^*$ & $\Delta_{\mathrm{frac}}$ & Base & $\alpha^*$ & $\Delta_{\mathrm{frac}}$ \\
\hline
\endfirsthead
\hline
\multirow{2}{*}{Subtask} & \multicolumn{3}{c|}{DS judge} & \multicolumn{3}{c|}{Llama 1B} & \multicolumn{3}{c|}{Llama 3B} & \multicolumn{3}{c|}{Llama 8B} & \multicolumn{3}{c|}{Qwen 8B} & \multicolumn{3}{c|}{Qwen 30B-A3B} \\
 & \textcolor{FKColor!70!black}{FK} & \textcolor{LUColor!85!black}{LU} & \textcolor{DRColor!85!black}{DR} & Base & $\alpha^*$ & $\Delta_{\mathrm{frac}}$ & Base & $\alpha^*$ & $\Delta_{\mathrm{frac}}$ & Base & $\alpha^*$ & $\Delta_{\mathrm{frac}}$ & Base & $\alpha^*$ & $\Delta_{\mathrm{frac}}$ & Base & $\alpha^*$ & $\Delta_{\mathrm{frac}}$ \\
\hline
\endhead
\hline
\multicolumn{19}{r}{Continued on next page}\\
\endfoot
\hline
\endlastfoot
\rowcolor{gray!14} agieval\_\allowbreak{aqua\_\allowbreak{}rat} & \cellcolor{FKColor!3!CapZero}0.034 & \cellcolor{LUColor!32!CapZero}0.322 & \cellcolor{DRColor!64!CapZero}0.644 & 0.248 & \textcolor{AlphaNeg!95!black}{-0.2} & +0.016 & 0.205 & \textcolor{AlphaNeg!80!black}{-0.15} & +0.028 & 0.232 & \textcolor{AlphaNeg!50!black}{-0.05} & +0.035 & 0.394 & \textcolor{AlphaPos!65!black}{+0.1} & +0.028 & 0.469 & \textcolor{AlphaPos!50!black}{+0.05} & +0.020 \\ \hline
agieval\_\allowbreak{gaokao\_\allowbreak{}biology} & \cellcolor{FKColor!68!CapZero}0.683 & \cellcolor{LUColor!15!CapZero}0.151 & \cellcolor{DRColor!17!CapZero}0.167 & 0.305 & \textcolor{AlphaNeg!80!black}{-0.15} & +0.019 & 0.405 & \textcolor{AlphaPos!65!black}{+0.1} & +0.014 & 0.490 & \textcolor{AlphaNeg!50!black}{-0.05} & +0.014 & 0.714 & \textcolor{AlphaNeg!50!black}{-0.05} & +0.029 & 0.795 & \textcolor{AlphaNeg!80!black}{-0.15} & +0.010 \\ \hline
\rowcolor{gray!14} agieval\_\allowbreak{gaokao\_\allowbreak{}chemistry} & \cellcolor{FKColor!69!CapZero}0.688 & \cellcolor{LUColor!8!CapZero}0.081 & \cellcolor{DRColor!23!CapZero}0.231 & 0.242 & \textcolor{AlphaPos!80!black}{+0.15} & +0.019 & 0.266 & \textcolor{AlphaPos!80!black}{+0.15} & +0.039 & 0.300 & \textcolor{AlphaPos!65!black}{+0.1} & +\,\textbf{0.058} & 0.444 & \textcolor{AlphaPos!50!black}{+0.05} & +0.024 & 0.502 & \textcolor{AlphaNeg!65!black}{-0.1} & +0.010 \\ \hline
agieval\_\allowbreak{gaokao\_\allowbreak{}chinese} & \cellcolor{FKColor!16!CapZero}0.156 & \cellcolor{LUColor!62!CapZero}0.617 & \cellcolor{DRColor!23!CapZero}0.228 & 0.224 & \textcolor{AlphaNeg!80!black}{-0.15} & +0.041 & 0.321 & \textcolor{AlphaPos!50!black}{+0.05} & +0.012 & 0.411 & \textcolor{AlphaZero!90!black}{0} & +0.000 & 0.557 & \textcolor{AlphaNeg!50!black}{-0.05} & +0.020 & 0.650 & \textcolor{AlphaPos!50!black}{+0.05} & +0.004 \\
\hline
\end{longtable}
\end{landscape}

\section{Layer-wise Non-uniformity of Alpha SVD Perturbations}
\label{sec:appendix-layerwise}

This appendix summarizes the spectral evidence that the same global $\alpha$ perturbation does not produce a uniform update strength across model depth. The reason is that each layer starts from a different singular-value distribution. For each MLP \texttt{up\_proj} and \texttt{down\_proj} matrix, the perturbation is applied as
\[
s_i' = G\left(\frac{s_i}{G}\right)^{1+\alpha},
\qquad
G = \exp\!\left(\frac{1}{n}\sum_i \log s_i\right),
\]
where here $s_i \equiv \sigma_i$ and $n=r$ relative to Sec.~\ref{sec:setup}; this is the same geometric-mean-preserving transform as Eqs.~(3)--(6), rewritten to emphasize the normalization. Positive $\alpha$ increases spectral inequality by pushing large singular values farther above the geometric mean and small singular values farther below it, whereas negative $\alpha$ smooths the spectrum. The sign of the perturbation is therefore global, but its realized magnitude is layer-specific.

\begin{figure*}[t]
\centering
\begin{minipage}[t]{0.49\textwidth}
\centering
\includegraphics[width=\linewidth,height=0.28\textheight,keepaspectratio]{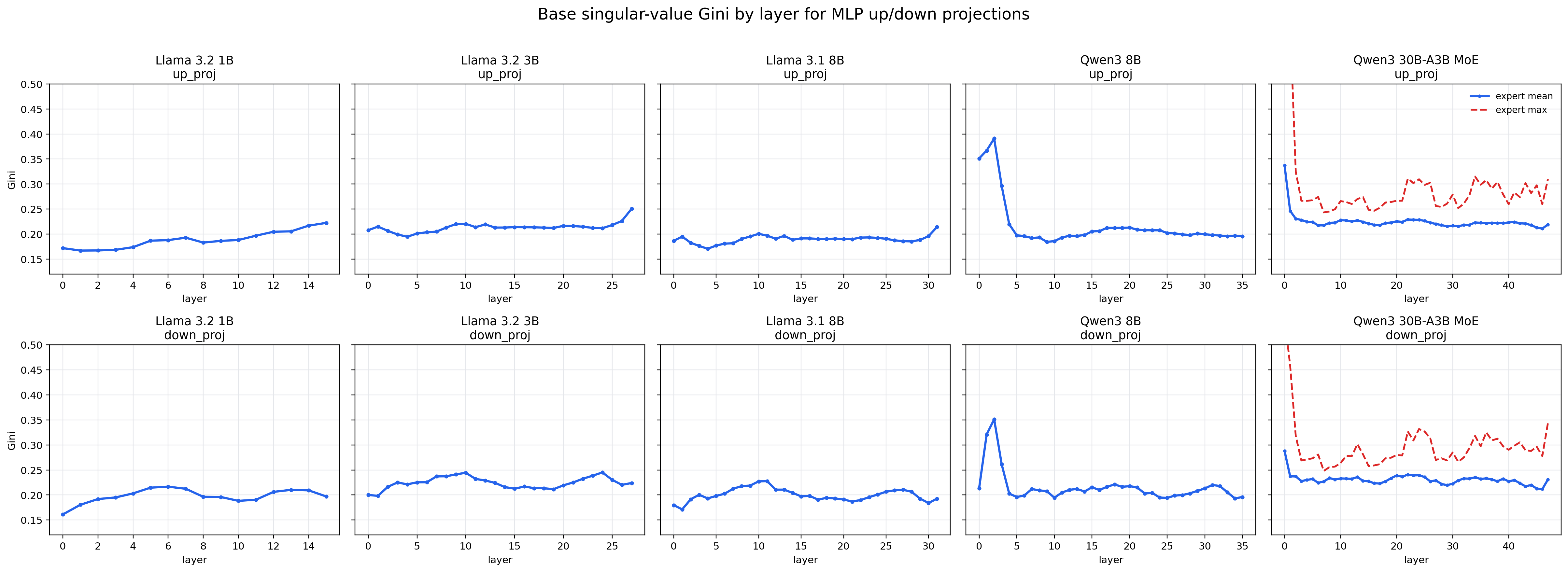}

\vspace{2pt}
\footnotesize\textbf{(a)} Layer-wise Gini coefficients.
\end{minipage}\hfill
\begin{minipage}[t]{0.49\textwidth}
\centering
\includegraphics[width=\linewidth,height=0.28\textheight,keepaspectratio]{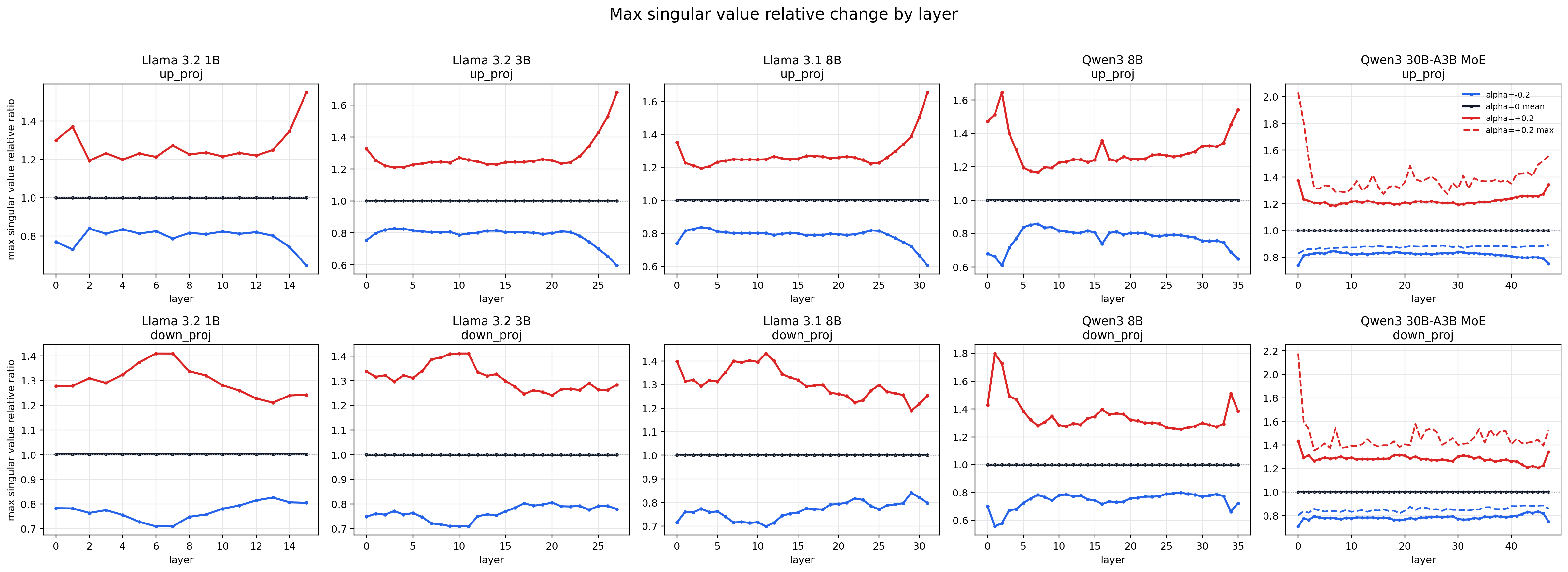}

\vspace{2pt}
\footnotesize\textbf{(b)} Relative change of the top singular value.
\end{minipage}
\caption{Two complementary views of layer-wise non-uniformity under the $\alpha$ SVD perturbation. Layers with more unequal spectra receive larger realized perturbations even when the global control parameter is held fixed.}
\label{fig:svd-layer-nonuniformity-main}
\end{figure*}

\begin{figure*}[t]
\centering
\begin{minipage}[t]{0.49\textwidth}
\centering
\includegraphics[width=\linewidth,height=0.28\textheight,keepaspectratio]{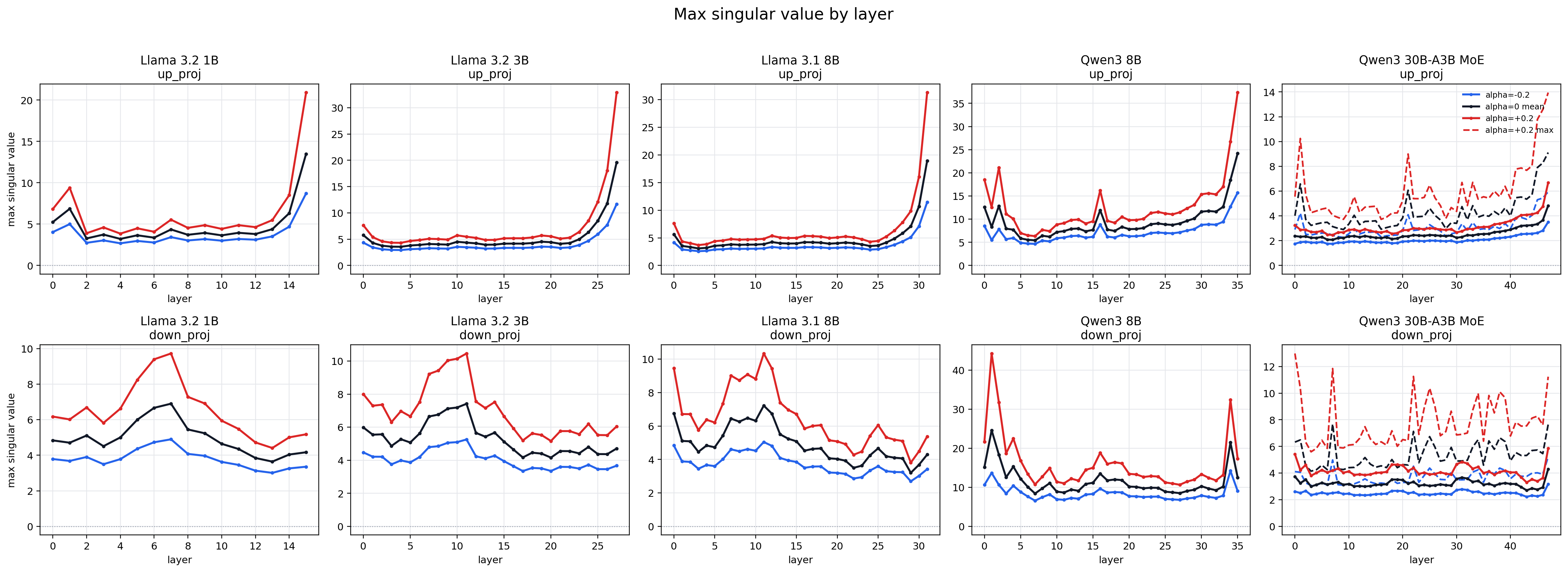}

\vspace{2pt}
\footnotesize\textbf{(a)} Absolute top singular values across layers.
\end{minipage}\hfill
\begin{minipage}[t]{0.49\textwidth}
\centering
\includegraphics[width=\linewidth,height=0.28\textheight,keepaspectratio]{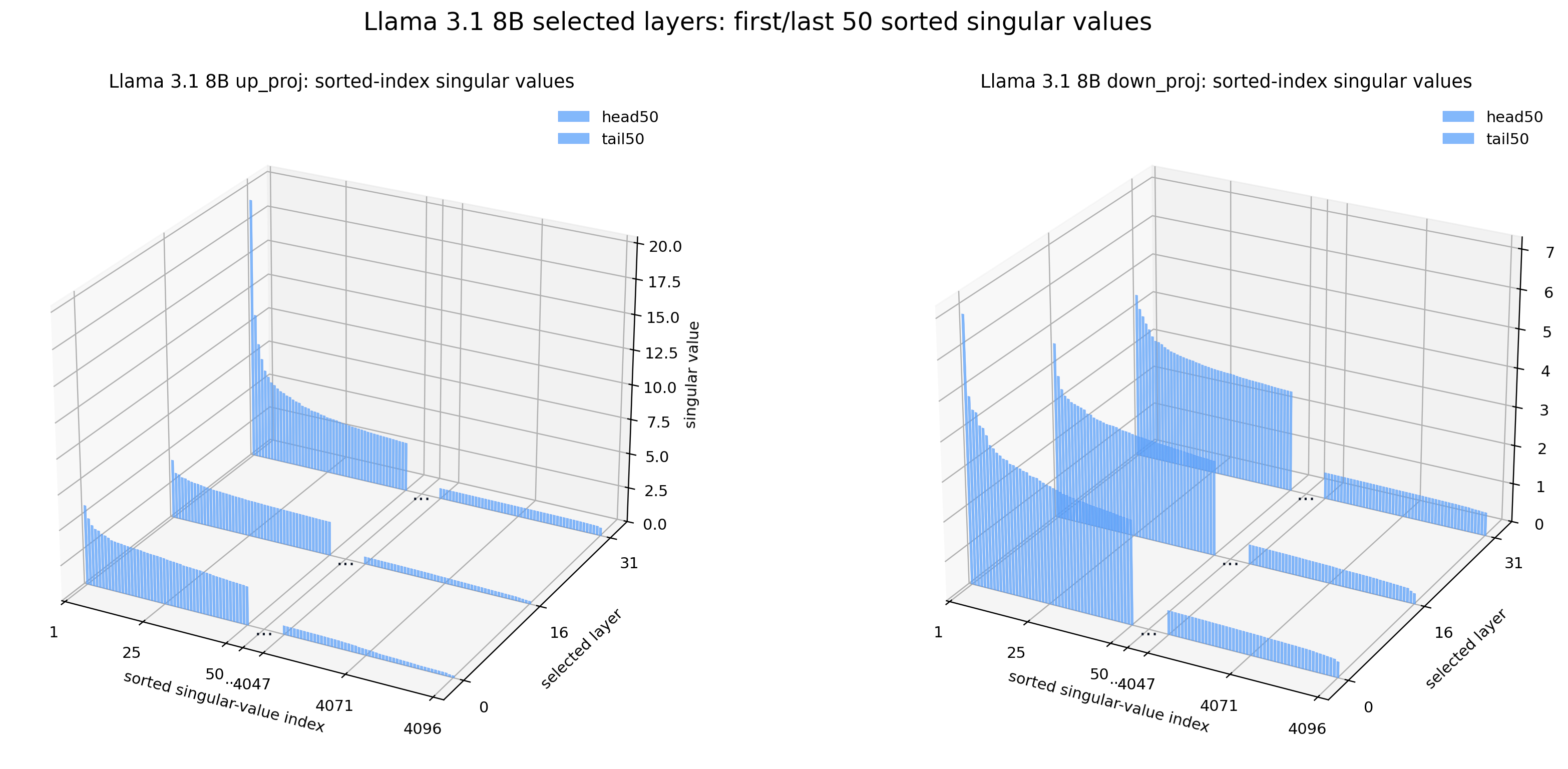}

\vspace{2pt}
\footnotesize\textbf{(b)} Head/tail singular values for selected Llama 3.1 8B layers.
\end{minipage}
\caption{Additional layer-wise diagnostics. The absolute-value plot exposes where already dominant directions can receive further amplification, while the Llama 3.1 8B visualization shows that the perturbation acts on layer-specific head and tail shapes rather than on a uniform spectrum.}
\label{fig:svd-layer-nonuniformity-secondary}
\end{figure*}

\paragraph{Layer-wise evidence.}
The Gini plot shows that spectral inequality differs substantially by layer, model, and projection. Dense Llama models exhibit strong depth dependence, especially in \texttt{up\_proj}; Qwen3 8B has unusually concentrated early layers; and Qwen3 30B-A3B MoE adds expert-level variation, where the maximum expert can deviate substantially from the layer mean. The relative top-singular-value plot identifies where the same global perturbation causes the largest multiplicative amplification or suppression, with the strongest effects occurring in layers whose top singular value is already far above the geometric mean.

\begin{table}[t]
\centering
\small
\begin{tabular}{llrr}
\toprule
Model / proj. & Layer & Base Gini & Top-SV multiplier \\
\midrule
Qwen3 8B / \texttt{down\_proj} & 1 & 0.3209 & 1.7984 \\
Qwen3 8B / \texttt{down\_proj} & 2 & 0.3513 & 1.7278 \\
Llama 3.2 3B Instruct / \texttt{up\_proj} & 27 & 0.2506 & 1.6784 \\
Llama 3.1 8B Instruct / \texttt{up\_proj} & 31 & 0.2143 & 1.6535 \\
Qwen3 8B / \texttt{up\_proj} & 2 & 0.3910 & 1.6443 \\
\bottomrule
\end{tabular}
\caption{Examples of large top-singular-value amplification at $\alpha=+0.2$. These cases illustrate that the same global perturbation can produce markedly different realized updates across layers.}
\label{tab:layerwise-top-sv-examples}
\end{table}

\paragraph{Interpretation.}
The data support three points. First, the $\alpha$ transform has a consistent sign effect: positive $\alpha$ increases Gini and top singular values, while negative $\alpha$ decreases them. Second, the magnitude of this effect is highly non-uniform across layers: layers with more unequal singular-value distributions are more affected by the same $\alpha$ than flatter layers, and the strongest affected layers differ across model families and projections. Third, MoE models add another axis of non-uniformity because the expert mean describes the typical routed expert while the expert maximum reveals that some individual experts are much more spectrally concentrated than the layer average.

\paragraph{Implication for experiments.}
Using one global $\alpha$ remains useful as a controlled intervention, but it should be interpreted as a spectrum-dependent perturbation rather than a uniform update. This explains why $\alpha$ sweeps can produce selective task improvements or degradations rather than smooth global quality changes. The layer-wise diagnostics motivate safer future variants such as layer-normalized $\alpha$, projection-specific $\alpha$ for \texttt{up\_proj} and \texttt{down\_proj}, MoE expert clipping, and reporting realized spectral statistics alongside the nominal $\alpha$ value.

\clearpage
\onecolumn

\section{Auxiliary-Task Alpha Sweeps Outside the Capability-Labeled Benchmark Suite}
\label{sec:appendix-aux-sweeps}

Table~\ref{tab:standard-judge-cot-all-alpha} records the official full non-judge auxiliary sweeps on the nine-point $\alpha$ grid from $-0.2$ to $+0.2$. It complements the condensed best-$\alpha$ summaries in Tables~\ref{tab:auxiliary-cross-eval} and \ref{tab:auxiliary-standout-diagnostics} by showing where gains are narrow, broad, or sharply non-monotone across the full grid. Judge-scored auxiliary probes are summarized in the main text but are kept separate from this official full non-judge sweep table.

\begin{landscape}
\scriptsize
\setlength{\tabcolsep}{2.2pt}
\begin{longtable}{llccccccccccc}
\caption{Official full non-judge auxiliary-task $\alpha$ sweeps. Each row reports all evaluated perturbation strengths from $-0.2$ to $+0.2$; the row-wise best score is bolded. $\alpha^*$ is the best $\alpha$ on the displayed grid, with ties assigned to $\alpha=0$ when the baseline is tied, and $\Delta_{\mathrm{frac}}$ is the gain over $\alpha=0$ as a fraction of the task's score range. BFCL uses retry-enabled success.}\\
\label{tab:standard-judge-cot-all-alpha}\\
\toprule
Task & Model & 
$-0.2$ & $-0.15$ & $-0.1$ & $-0.05$ & $0$ & $+0.05$ & $+0.1$ & $+0.15$ & $+0.2$ & $\alpha^*$ & $\Delta_{\mathrm{frac}}$ \\
\midrule
\endfirsthead
\toprule
Task & Model & $-0.2$ & $-0.15$ & $-0.1$ & $-0.05$ & $0$ & $+0.05$ & $+0.1$ & $+0.15$ & $+0.2$ & $\alpha^*$ & $\Delta_{\mathrm{frac}}$ \\
\midrule
\endhead
\multicolumn{13}{l}{\textit{Standard no-CoT}}\\

GSM8K & Llama-3.2-1B & $0.213$ & $\mathbf{0.223}$ & $0.209$ & $0.172$ & $0.144$ & $0.114$ & $0.080$ & $0.061$ & $0.033$ & $-0.15$ & $+0.079$ \\
 & Llama-3.2-3B & $0.121$ & $\mathbf{0.154}$ & $0.107$ & $0.112$ & $0.121$ & $0.123$ & $0.122$ & $0.121$ & $0.125$ & $-0.15$ & $+0.033$ \\
 & Llama-3.1-8B & $0.248$ & $\mathbf{0.551}$ & $0.506$ & $0.416$ & $0.324$ & $0.305$ & $0.375$ & $0.250$ & $0.159$ & $-0.15$ & $+0.227$ \\
 & Qwen3-8B & $0.798$ & $0.843$ & $\mathbf{0.851}$ & $0.845$ & $0.845$ & $0.832$ & $0.809$ & $0.763$ & $0.683$ & $-0.10$ & $+0.006$ \\
 & Qwen3-30B-A3B & $0.858$ & $0.839$ & $0.861$ & $0.867$ & $0.864$ & $0.879$ & $\mathbf{0.898}$ & $0.886$ & $0.888$ & $+0.10$ & $+0.033$ \\
\addlinespace[1pt]

MATH & Llama-3.2-1B & $0.138$ & $\mathbf{0.152}$ & $0.150$ & $0.140$ & $0.138$ & $\mathbf{0.152}$ & $0.136$ & $0.100$ & $0.086$ & $-0.15$ & $+0.014$ \\
 & Llama-3.2-3B & $0.140$ & $0.170$ & $\mathbf{0.204}$ & $0.194$ & $0.178$ & $0.198$ & $0.194$ & $0.186$ & $0.154$ & $-0.10$ & $+0.026$ \\
 & Llama-3.1-8B & $0.158$ & $0.222$ & $0.220$ & $0.218$ & $0.212$ & $0.208$ & $\mathbf{0.228}$ & $0.216$ & $0.154$ & $+0.10$ & $+0.016$ \\
 & Qwen3-8B & $\mathbf{0.394}$ & $\mathbf{0.394}$ & $0.374$ & $0.376$ & $0.350$ & $0.336$ & $0.318$ & $0.300$ & $0.292$ & $-0.20$ & $+0.044$ \\
 & Qwen3-30B-A3B & $0.440$ & $0.420$ & $0.426$ & $0.424$ & $0.426$ & $0.432$ & $0.428$ & $\mathbf{0.448}$ & $0.414$ & $+0.15$ & $+0.022$ \\
\addlinespace[1pt]

ARC-C & Llama-3.2-1B & $0.385$ & $0.443$ & $0.474$ & $0.494$ & $0.497$ & $\mathbf{0.498}$ & $0.491$ & $0.445$ & $0.352$ & $+0.05$ & $+0.001$ \\
 & Llama-3.2-3B & $0.387$ & $0.664$ & $0.712$ & $0.724$ & $0.728$ & $\mathbf{0.732}$ & $0.724$ & $0.693$ & $0.651$ & $+0.05$ & $+0.004$ \\
 & Llama-3.1-8B & $0.706$ & $0.789$ & $0.811$ & $\mathbf{0.820}$ & $0.815$ & $0.813$ & $0.801$ & $0.779$ & $0.731$ & $-0.05$ & $+0.005$ \\
 & Qwen3-8B & $0.907$ & $0.910$ & $0.913$ & $0.916$ & $\mathbf{0.922}$ & $0.914$ & $0.907$ & $0.897$ & $0.873$ & $0$ & $+0.000$ \\
 & Qwen3-30B-A3B & $0.924$ & $0.925$ & $0.925$ & $0.926$ & $\mathbf{0.927}$ & $0.925$ & $0.922$ & $0.920$ & $0.924$ & $0$ & $+0.000$ \\
\addlinespace[1pt]

BFCL & Llama-3.2-1B & $0.637$ & $0.738$ & $\mathbf{0.743}$ & $0.718$ & $0.598$ & $0.440$ & $0.340$ & $0.275$ & $0.225$ & $-0.10$ & $+0.145$ \\
 & Llama-3.2-3B & $\mathbf{0.172}$ & $0.142$ & $0.113$ & $0.035$ & $0.048$ & $0.128$ & $0.092$ & $0.140$ & $0.138$ & $-0.20$ & $+0.125$ \\
 & Llama-3.1-8B & $0.355$ & $0.383$ & $0.417$ & $0.500$ & $\mathbf{0.593}$ & $0.512$ & $0.295$ & $0.177$ & $0.177$ & $0$ & $+0.000$ \\
 & Qwen3-8B & $0.772$ & $\mathbf{0.805}$ & $0.800$ & $0.755$ & $0.777$ & $0.728$ & $0.675$ & $0.575$ & $0.640$ & $-0.15$ & $+0.028$ \\
 & Qwen3-30B-A3B & $0.905$ & $0.905$ & $\mathbf{0.912}$ & $\mathbf{0.912}$ & $0.885$ & $0.853$ & $0.782$ & $0.738$ & $0.703$ & $-0.10$ & $+0.027$ \\
\addlinespace[1pt]

\multicolumn{13}{l}{\textit{Standard CoT}}\\

GSM8K-CoT & Llama-3.2-1B & $0.284$ & $0.315$ & $0.349$ & $0.357$ & $\mathbf{0.376}$ & $0.359$ & $0.361$ & $0.303$ & $0.230$ & $0$ & $+0.000$ \\
 & Llama-3.2-3B & $0.458$ & $0.635$ & $0.691$ & $\mathbf{0.747}$ & $0.745$ & $0.736$ & $0.718$ & $0.682$ & $0.616$ & $-0.05$ & $+0.002$ \\
 & Llama-3.1-8B & $0.588$ & $0.752$ & $0.820$ & $\mathbf{0.837}$ & $0.807$ & $0.788$ & $0.792$ & $0.757$ & $0.654$ & $-0.05$ & $+0.030$ \\
 & Qwen3-8B & $0.331$ & $\mathbf{0.363}$ & $0.318$ & $0.322$ & $0.306$ & $0.296$ & $0.279$ & $0.271$ & $0.224$ & $-0.15$ & $+0.057$ \\
 & Qwen3-30B-A3B & $0.500$ & $\mathbf{0.514}$ & $0.508$ & $0.511$ & $0.469$ & $0.455$ & $0.444$ & $0.397$ & $0.373$ & $-0.15$ & $+0.045$ \\
\addlinespace[1pt]

MATH-CoT & Llama-3.2-1B & $0.206$ & $0.256$ & $0.246$ & $0.254$ & $0.280$ & $0.260$ & $\mathbf{0.284}$ & $0.236$ & $0.180$ & $+0.10$ & $+0.004$ \\
 & Llama-3.2-3B & $0.220$ & $0.370$ & $0.410$ & $0.420$ & $\mathbf{0.464}$ & $0.454$ & $0.398$ & $0.420$ & $0.290$ & $0$ & $+0.000$ \\
 & Llama-3.1-8B & $0.174$ & $0.354$ & $0.412$ & $0.468$ & $\mathbf{0.504}$ & $0.458$ & $0.434$ & $0.384$ & $0.262$ & $0$ & $+0.000$ \\
 & Qwen3-8B & $0.194$ & $0.166$ & $0.164$ & $\mathbf{0.202}$ & $0.172$ & $0.172$ & $0.154$ & $0.136$ & $0.152$ & $-0.05$ & $+0.030$ \\
 & Qwen3-30B-A3B & $\mathbf{0.208}$ & $0.186$ & $0.198$ & $0.152$ & $0.178$ & $0.184$ & $0.168$ & $0.164$ & $0.172$ & $-0.20$ & $+0.030$ \\
\addlinespace[1pt]

ARC-C-CoT & Llama-3.2-1B & $0.084$ & $0.083$ & $0.149$ & $0.201$ & $0.208$ & $\mathbf{0.226}$ & $0.216$ & $0.217$ & $0.159$ & $+0.05$ & $+0.018$ \\
 & Llama-3.2-3B & $0.307$ & $0.510$ & $0.608$ & $0.654$ & $0.663$ & $0.671$ & $\mathbf{0.673}$ & $0.665$ & $0.556$ & $+0.10$ & $+0.010$ \\
 & Llama-3.1-8B & $0.718$ & $0.817$ & $0.834$ & $\mathbf{0.840}$ & $0.839$ & $0.831$ & $0.816$ & $0.794$ & $0.730$ & $-0.05$ & $+0.002$ \\
 & Qwen3-8B & $0.506$ & $0.551$ & $0.555$ & $0.558$ & $0.583$ & $0.606$ & $\mathbf{0.614}$ & $0.596$ & $0.543$ & $+0.10$ & $+0.032$ \\
 & Qwen3-30B-A3B & $0.778$ & $0.794$ & $0.783$ & $0.799$ & $0.819$ & $0.822$ & $\mathbf{0.828}$ & $0.823$ & $0.816$ & $+0.10$ & $+0.009$ \\
\addlinespace[1pt]

\bottomrule
\end{longtable}
\end{landscape}

\bibliographystyle{plainnat}
\bibliography{refs}

@misc{svd_transformer_interpretable,
  title        = {The Singular Value Decompositions of Transformer Weight Matrices are Highly Interpretable},
  author       = {{Conjecture}},
  year         = {2022},
  month        = nov,
  day          = {28},
  howpublished = {Technical blog post},
  url          = {https://www.conjecture.dev/research/the-singular-value-decompositions-of-transformer-weight-matrices-are-highly-interpretable},
  note         = {Accessed 2026-05-07}
}

@misc{beyond_components_svd_circuits,
  title         = {Beyond Components: Singular Vector-Based Interpretability of Transformer Circuits},
  author        = {Ahmad, Areeb and Joshi, Abhinav and Modi, Ashutosh},
  year          = {2025},
  eprint        = {2511.20273},
  archivePrefix = {arXiv},
  primaryClass  = {cs.LG},
  url           = {https://arxiv.org/abs/2511.20273}
}

@misc{spectral_filters_dark_signals,
  title         = {Spectral Filters, Dark Signals, and Attention Sinks},
  author        = {Cancedda, Nicola},
  year          = {2024},
  eprint        = {2402.09221},
  archivePrefix = {arXiv},
  primaryClass  = {cs.AI},
  url           = {https://arxiv.org/abs/2402.09221}
}

@misc{small_singular_values_matter,
  title         = {Small Singular Values Matter: A Random Matrix Analysis of Transformer Models},
  author        = {Staats, Max and Thamm, Matthias and Rosenow, Bernd},
  year          = {2025},
  eprint        = {2410.17770},
  archivePrefix = {arXiv},
  primaryClass  = {cs.LG},
  url           = {https://arxiv.org/abs/2410.17770}
}

@misc{laser,
  title         = {The Truth is in There: Improving Reasoning in Language Models with Layer-Selective Rank Reduction},
  author        = {Sharma, Pratyusha and Ash, Jordan T. and Misra, Dipendra},
  year          = {2023},
  eprint        = {2312.13558},
  archivePrefix = {arXiv},
  primaryClass  = {cs.LG},
  url           = {https://arxiv.org/abs/2312.13558}
}

@misc{pissa,
  title         = {{PiSSA}: Principal Singular Values and Singular Vectors Adaptation of Large Language Models},
  author        = {Meng, Fanxu and Wang, Zhaohui and Zhang, Muhan},
  year          = {2025},
  eprint        = {2404.02948},
  archivePrefix = {arXiv},
  primaryClass  = {cs.LG},
  url           = {https://arxiv.org/abs/2404.02948}
}

@misc{milora,
  title         = {{MiLoRA}: Harnessing Minor Singular Components for Parameter-Efficient {LLM} Finetuning},
  author        = {Wang, Hanqing and Li, Yixia and Wang, Shuo and Chen, Guanhua and Chen, Yun},
  year          = {2025},
  eprint        = {2406.09044},
  archivePrefix = {arXiv},
  primaryClass  = {cs.CL},
  url           = {https://arxiv.org/abs/2406.09044}
}

@misc{kasa,
  title         = {{KaSA}: Knowledge-Aware Singular-Value Adaptation of Large Language Models},
  author        = {Wang, Fan and Jiang, Juyong and Park, Chansung and Kim, Sunghun and Tang, Jing},
  year          = {2025},
  eprint        = {2412.06071},
  archivePrefix = {arXiv},
  primaryClass  = {cs.CL},
  url           = {https://arxiv.org/abs/2412.06071}
}

@misc{geva_kv_memory,
  title         = {Transformer Feed-Forward Layers Are Key-Value Memories},
  author        = {Geva, Mor and Schuster, Roei and Berant, Jonathan and Levy, Omer},
  year          = {2021},
  eprint        = {2012.14913},
  archivePrefix = {arXiv},
  primaryClass  = {cs.CL},
  url           = {https://arxiv.org/abs/2012.14913}
}

@misc{rome,
  title         = {Locating and Editing Factual Associations in {GPT}},
  author        = {Meng, Kevin and Bau, David and Andonian, Alex and Belinkov, Yonatan},
  year          = {2023},
  eprint        = {2202.05262},
  archivePrefix = {arXiv},
  primaryClass  = {cs.CL},
  url           = {https://arxiv.org/abs/2202.05262}
}

@misc{memit,
  title         = {Mass-Editing Memory in a Transformer},
  author        = {Meng, Kevin and Sharma, Arnab Sen and Andonian, Alex and Belinkov, Yonatan and Bau, David},
  year          = {2023},
  eprint        = {2210.07229},
  archivePrefix = {arXiv},
  primaryClass  = {cs.CL},
  url           = {https://arxiv.org/abs/2210.07229}
}

@misc{knowledge_neurons,
  title         = {Knowledge Neurons in Pretrained Transformers},
  author        = {Dai, Damai and Dong, Li and Hao, Yaru and Sui, Zhifang and Chang, Baobao and Wei, Furu},
  year          = {2022},
  eprint        = {2104.08696},
  archivePrefix = {arXiv},
  primaryClass  = {cs.CL},
  url           = {https://arxiv.org/abs/2104.08696}
}

@misc{llama3_herd,
  title         = {The Llama 3 Herd of Models},
  author        = {Grattafiori, Aaron and Dubey, Abhimanyu and Jauhri, Abhinav and Pandey, Abhinav and Kadian, Abhishek and Al-Dahle, Ahmad and Letman, Aiesha and Mathur, Akhil and Schelten, Alan and Vaughan, Alex and others},
  year          = {2024},
  eprint        = {2407.21783},
  archivePrefix = {arXiv},
  primaryClass  = {cs.AI},
  url           = {https://arxiv.org/abs/2407.21783}
}

@misc{qwen3,
  title         = {{Qwen3} Technical Report},
  author        = {Yang, An and Li, Anfeng and Yang, Baosong and Zhang, Beichen and Hui, Binyuan and Zheng, Bo and Yu, Bowen and Gao, Chang and others},
  year          = {2025},
  eprint        = {2505.09388},
  archivePrefix = {arXiv},
  primaryClass  = {cs.CL},
  url           = {https://arxiv.org/abs/2505.09388}
}

@misc{trait,
  title         = {Do {LLMs} Have Distinct and Consistent Personality? {TRAIT}: Personality Testset Designed for {LLMs} with Psychometrics},
  author        = {Lee, Seungbeen and Lim, Seungwon and Han, Seungju and Oh, Giyeong and Chae, Hyungjoo and Chung, Jiwan and Kim, Minju and Kwak, Beong-woo and Lee, Yeonsoo and Lee, Dongha and others},
  year          = {2024},
  eprint        = {2406.14703},
  archivePrefix = {arXiv},
  primaryClass  = {cs.CL},
  url           = {https://arxiv.org/abs/2406.14703}
}

\end{document}